\documentclass[final]{cvpr}
\usepackage{times}
\usepackage{epsfig}
\usepackage{graphicx}
\usepackage{amsmath}
\usepackage{amssymb}
\usepackage{subcaption}

\usepackage[pagebackref=true,breaklinks=true,colorlinks,bookmarks=false]{hyperref}

\def\cvprPaperID{10440} %
\def\confYear{CVPR 2021}

\begin{document}

\title{Black-box Explanation of Object Detectors via Saliency Maps} %
\author{
Vitali Petsiuk\thanks{Work completed while an intern at Adobe Research.\newline 
This work was partially supported by the DARPA XAI program.\newline
Project page: \href{https://cs-people.bu.edu/vpetsiuk/drise/}{https://cs-people.bu.edu/vpetsiuk/drise/}}\\
Boston University\\
{\tt\small vpetsiuk@bu.edu}
\and
Rajiv Jain\\
Adobe Research\\
{\tt\small rajijain@adobe.com}
\and
Varun Manjunatha\\
Adobe Research\\
{\tt\small vmanjuna@adobe.com}
\and
Vlad I. Morariu\\
Adobe Research\\
{\tt\small morariu@adobe.com}
\and
Ashutosh Mehra\\
Adobe Document Cloud\\
{\tt\small amehra@adobe.com}
\and
Vicente Ordonez\\
University of Virginia\\
{\tt\small vicente@virginia.edu}
\and
Kate Saenko\\
Boston University,\\ MIT-IBM Watson AI Lab\\
{\tt\small saenko@bu.edu}
}

\def\eg{\emph{e.g.}} \def\Eg{\emph{E.g.}}
\def\ie{\emph{i.e.}} \def\Ie{\emph{I.e.}}

\def\methodname{D-RISE}
\def\fullmethodname{\textit{Detector Randomized Input Sampling for Explanation}}

\maketitle

\begin{abstract}
    We propose \methodname{}, a method for generating visual explanations for the predictions of object detectors. 
    Utilizing the proposed similarity metric that accounts for both localization and categorization aspects of object detection allows our method to produce saliency maps that show image areas that most affect the prediction. 
    \methodname{} can be considered ``black-box'' in the software testing sense, as it only needs access to the inputs and outputs of an object detector. Compared to gradient-based methods, \methodname{} is more general and agnostic to the particular type of object detector being tested, and does not need knowledge of the inner workings of the model.
    We show that \methodname{} can be easily applied to different object detectors including one-stage detectors such as YOLOv3 and two-stage detectors such as Faster-RCNN. We present a detailed analysis of the generated visual explanations to highlight the utilization of context and possible biases learned by object detectors. %
\end{abstract}

\section{Introduction}
\newcommand{\wfour}{0.24\linewidth}

\begin{figure}[t]
    \centering
    
    \includegraphics[width=\wfour]{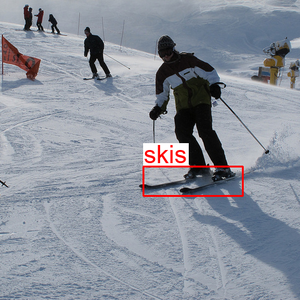}
    \includegraphics[width=\wfour]{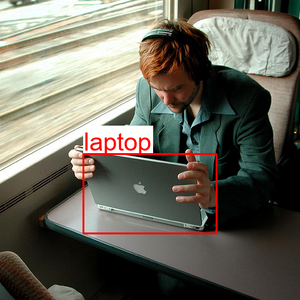}
    \includegraphics[width=\wfour]{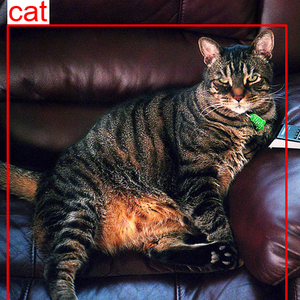}
    \includegraphics[width=\wfour]{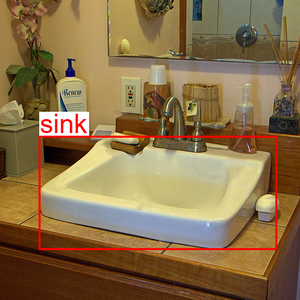}
    
    \includegraphics[width=\wfour]{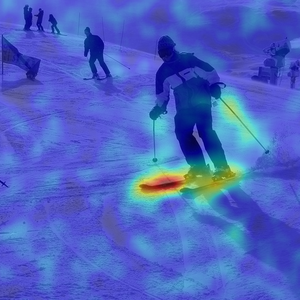}
    \includegraphics[width=\wfour]{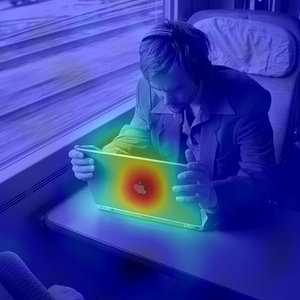}
    \includegraphics[width=\wfour]{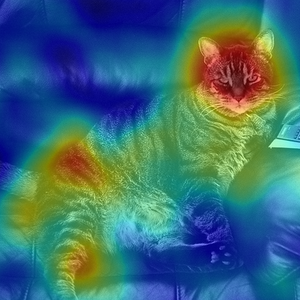}
    \includegraphics[width=\wfour]{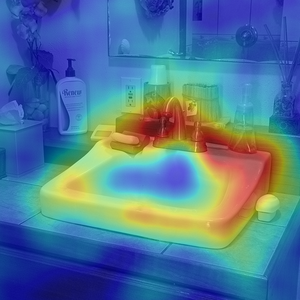}
    
    \caption{\small{\methodname{} can highlight which regions of an image were used by an object detector. Here we show outputs for a few corresponding images where importance increases from blue to red. In these examples, \methodname{} reveals things such as detectors often looking outside bounding boxes to detect objects e.g., looking at the ski poles to predict skis, or looking to a subset of regions within the object e.g., looking at the Apple logo to predict laptop.}}
    \label{fig:misalignment}
    \vspace{-0.4cm}
\end{figure}
\begin{figure*}
    \begin{center}
        \includegraphics[width=0.9\linewidth]{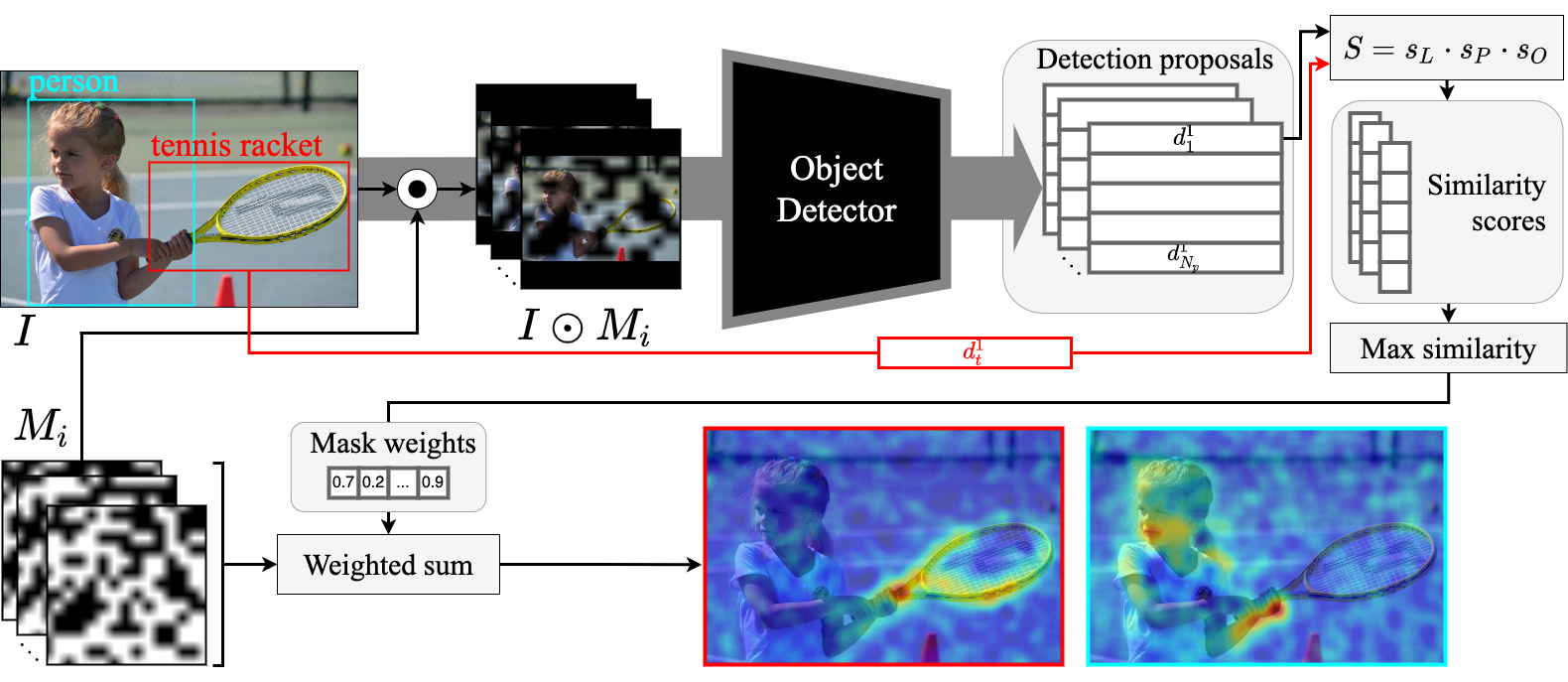}
    \end{center}
    \caption{\small{Our method \methodname{} attempts to explain the detections (bounding-box+category) produced for this image by an object detector. We convert target detections that need to be explained into detection vectors $d_t$. We sample $N$ binary masks, $M_i$, and run the detector on the masked images to obtain proposals $D_p$. We compute pairwise similarities between targets and proposals to obtain weights for each mask. Finally, the weighted sum of masks is computed to produce saliency maps. In classification, the ouput of the black-box model can be directly used as mask weights.}}
    \label{fig:overview}
    \vspace{-0.4cm}
\end{figure*}
The field of object detection has experienced significant gains in performance since the adoption of deep neural networks (DNNs)~\cite{girshick14rcnn}. %
However, DNNs remain opaque tools with a complex and unintuitive process of decision-making, resulting in them being hard to understand, debug and improve. 
A number of different explanation techniques offer potential solutions to these issues. 
They have already been shown to find biases in trained models~\cite{wang15cfrl}, help debug them \cite{koh17influence} and increase user's trust \cite{selvaraju17gradcam}.
A popular approach to explanation involves the use of attribution techniques which produce saliency maps~\cite{morch95saliency, simonyan13deepinside}, \ie, heatmaps representing the influence different pixels have on the model's decision. 
Hitherto, these techniques have primarily focused on the image classification task~\cite{petsiuk18rise, fong17mask, selvaraju17gradcam, Wang_2019_ICCV,zhou16CAM, bach15lrp, zeiler14viscnn}, with few addressing other problems such as visual question answering~\cite{park18vqasaliency}, video captioning~\cite{ramanishka17topdown, bargal18exbprnn} and video activity recognition~\cite{bargal18exbprnn}.
In this work, we address the relatively underexplored direction of generating saliency maps for object detectors.

Unlike methods that explain the emerging patterns in the learned weights or activations~\cite{bau17networkdis, zeiler14viscnn, olah2017feature}, attribution techniques are usually tightly connected to the model's design and they rely on a number of assumptions about the model's architecture. For example, Grad-CAM~\cite{selvaraju17gradcam} assumes that each feature map correlates with some concept, and therefore, feature maps can be weighted with respect to the importance of their concept for the output category. We show that these assumptions might not hold for object detection models, resulting in failure to produce quality saliency maps.
Additionally, object detectors require explanations not just for the categorization of a bounding box but also for the location of the bounding box itself. For these reasons, direct application of existing attribution techniques to object detectors is infeasible.

We propose \fullmethodname{}, or \methodname{}, the first method to produce saliency maps for object detectors that is capable of explaining both the localization and classification aspects of the detection. \methodname{} uses an input masking technique first proposed by RISE~\cite{petsiuk18rise}, which enables explanation of the more complex detection networks because it does not rely on gradients or the inner workings of the underlying object detector. However, the method in~\cite{petsiuk18rise} is only applicable to classification, not detection. \methodname{} is a black-box method and can be in principle applied to any object detector.

Explaining visual classifiers with saliency maps has allowed researchers to investigate the localization abilities implicitly learned by these models.
Moreover, some works have used explanations of visual classifiers for weakly-supervised object localization~\cite{lai17wsod, OquabBLS15}.
In object detection, however, the localization decisions of the model are explicit as they are expressed directly in the outputs of the model. 
Therefore, one might assume that exploring spatial importance in this case is redundant, and that the model has already predicted bounding boxes around everything it deems important. In our experiments with \methodname{}{}, we observe that DNN based object detectors also learn to utilize contextual regions outside of the box to detect objects. For instance the last column in Fig.~\ref{fig:misalignment} shows how the tap helps to localize the sink even when it is clearly outside the detected box. In fact, the importance of contextual information for object detection has long been established for both humans~\cite{biederman1982scene, oliva2007role} and machines~\cite{torralba03contextual, mottaghi14role}. Another reason for studying an object detector's saliency is the fact that not all sub-regions within the object's bounding box are equally important.
Some object parts are more discriminant, while others may occur with objects of different categories, \eg, cat faces are highlighted as more important by the network than its body (Fig.~\ref{fig:misalignment}).

Our contributions can be summarized as follows:
\begin{itemize}
    \itemsep 0em 
    \item We propose \methodname{}, a black-box attribution technique for explaining object detectors via saliency maps, by defining a detection similarity metric. 
    \item We demonstrate generalizability of \methodname{} by explaining two commonly used object detectors with different architectural designs, namely one-stage YOLOv3~\cite{redmon18yolov3} and two-stage Faster R-CNN~\cite{ren17fasterrcnn}.
    \item Using \methodname{}, we systematically analyze potential sources of errors and bias in commonly used object detectors trained on the MS-COCO~\cite{lin14coco} dataset and discover common patterns in data learned by the model.
    \item We evaluate our method using automated metrics from classification saliency and a user study. Additionally, we propose an evaluation procedure that measures how well the saliency method can discover deliberately introduced biases in the model via synthetic markers. Our method surpasses the classification baselines.
\end{itemize}

\section{Related Work}

\subsection{Object Detection}
Object detectors can be divided into two groups: two-stage detectors, with the Faster R-CNN~\cite{ren17fasterrcnn} being the most representative, and one-stage detectors, such as YOLO~\cite{redmon18yolov3}, SSD~\cite{liu16ssd} and CornerNet~\cite{law18cornernet}.
Two-stage detectors consist of a region proposal stage, where a sparse set of regions of interest (ROI) is selected, followed by a feature extraction stage for the subsequent classification of each candidate ROI. One-stage methods do not perform ROI pooling and instead use a single network to detect objects.
Our saliency technique is able to analyze both two-stage and single-stage detectors (Faster R-CNN and YOLO, respectively).

Previous works on explaining DNN-based object detectors include their feature space visualization~\cite{vondrick16visualizing}, analyzing the biases, such as pedestrians' skin color~\cite{Wilson19inequity}.
Two recent works have explored explainability using saliency for SSD object detectors \cite{tsunakawa19crp, gudovskiy18ex2fix}. These methods rely on tailored white-box approaches in comparison to \methodname{}, which treats detectors as black boxes.

\subsection{Visual Saliency Methods}
Various visual grounding techniques retrospectively provide explainability to computer vision models after they have been trained. 
Several saliency methods for classification models have been proposed. A first group of methods backpropagate an importance score through the layers of the neural network from the model's output to the individual pixels in the input, e.g. Gradients~\cite{simonyan13deepinside}, Excitation Backprop~\cite{zhang18exbp} and Layer-wise Relevance Propagation~\cite{bach15lrp}.
These methods are sometimes tailored to the model's architecture, \ie\ they cannot be used for new network architectures without implementing the propagation scheme for new layers. 
Grad-CAM~\cite{selvaraju17gradcam}, a generalization of CAM~\cite{zhou16CAM}, computes the regular gradients up to a selected intermediate layer and then combines them with the corresponding activations to get a low-resolution saliency map. While, a recent version of Grad-CAM has been proposed to study adversarial context patches in single-shot object detectors~\cite{Saha_2020_CVPR_Workshops}, most work has been on explaining classification results.

A second group of works perform specific perturbations on image regions, such as occlusion, adding noise, inpainting, and blurring.
After observing the effect of a perturbation on the model's output, the methods determine the importance of the region that was perturbed~\cite{simonyan13deepinside, petsiuk18rise, ribeiro16lime, LundbergL17SHAP, fong18mask, dabkowski17realtime}.
Specifically, Occlusion~\cite{zeiler14viscnn} blocks out square parts of the image in a sliding window manner and captures the drop in the class score to determine the importance.
LIME~\cite{ribeiro16lime} approximates the deep model by a linear classifier, trains it in the vicinity of the input point by using samples with occluded superpixels and uses the learned weights as the measure of superpixel importance. The Meaningful Perturbation approach~\cite{fong17mask} and Real Time Image Saliency~\cite{dabkowski17realtime} optimize the perturbation mask using gradient descent.\footnote{
    These two works also define themselves as black-box methods, however their definition of ``black-box'' is different from ours.
    While their methods can be applied to any differentiable image classification network, they still require access to model's weights and gradients for gradient descent optimization.
    Along with~\cite{zeiler14viscnn, ribeiro16lime, petsiuk18rise}, our work uses a stricter definition of ``black-box'', entirely prohibiting access to any of the model's internal parameters.
}
Finally, RISE~\cite{petsiuk18rise} generates a set of random masks, applies the classifier to masked versions of the input and uses the predicted class probabilities as weights, computing a weighted sum of the masks as the saliency map.  We use this masking technique to explain object detectors rather than image classifiers. Since this and other saliency methods described above cannot be directly applied to the detection task, our work extends the prior state of the art by enabling detector explanation. We describe key differences that have to be addressed for extending saliency methods to object detection in the next section.

Black-box and white-box approaches have slightly different use cases. Black-box methods, while typically slower at run time, can save developer time due to their higher generalizability and ease of application. They also enable analysis of proprietary models or APIs which cannot be studied using white-box approaches. Arguably, black-box methods are more intuitive, because they directly measure the effect that input ablations have on the model, without relying on heuristics such as rules for importance back-propagation. On the other hand, faster white-box methods are more suitable for large scale and real-time applications.

Methods above can only explain a scalar value in the model's output. In case of image classification, it is the class probability score that is used to either backpropagate from (gradient methods) or to gauge the effect of image modification (perturbation methods). 
In addition to class probabilities, an object detector has to predict the bounding box location of an object, and that also requires explanations. Moreover, it produces multiple detection proposals per image, which can greatly differ for modified versions of the image. These distinctions make the direct application of existing classification saliency methods infeasible. We show that gradient based methods do not produce quality saliency maps if used to explain a probability score in one of the detection vectors. In our work we address these issues, making it possible to produce saliency maps for object detectors.
\section{Method}

Given an $h$-by-$w$ image $I$, a DNN detector model $f$, and an object detection $d$ specified by a bounding box and a category label, our goal is to produce a saliency map $S$ to explain the detection. The map consists of $h$-by-$w$ values indicating the importance of each pixel in $I$ in influencing $f$ to predict $d$.
We propose \methodname{} to solve this problem in a black-box manner, \textit{i.e.}, without access to $f$'s weights, gradients or architecture.
Our method is inspired by the randomized perturbations (masks) applied to the image by the RISE model to explain object classifiers, except that we leverage the random-masking idea to explain object detectors. The main idea is to measure the effect of masking  randomized regions on the predicted output, using changes in $f$'s output to determine the importance.  Figure~\ref{fig:overview} shows an overview of our approach.

Existing approaches for image classification saliency cannot be directly applied to the object detection task. They assume a single categorical model output, while object detectors produce a multitude of detection vectors that encode class probabilities, localization information and additional information such as an objectness score. To apply random masking to detectors, we incorporate localization and objectness scores into the process of generating detector saliency maps.

Most detector networks, including Faster R-CNN and YOLO, produce a large number of bounding box proposals which are subsequently refined using confidence thresholding and non-maximum suppression to leave a small number of final detections.
We denote such bounding box proposals in the following manner:
\begin{align}
    d_i &= \big[L_i, O_i, P_i\big] \\ 
    &= \big[(x_1^i, y_1^i, x_2^i, y_2^i), O_i, (p_1^i, \ldots, p_C^i)\big]
\end{align}
Each proposal is encoded into a detection vector $d_i$ consisting of
\begin{itemize}
    \itemsep 0em
    \item localization information $L_i$, defining bounding box corners $(x_1^i, y_1^i)$ and $(x_2^i, y_2^i)$,
    \item objectness score $O_i\in [0, 1]$, representing the probability that bounding box $L_i$ contains an object of any class (if the detector does not produce such a score this term may be ignored), and
    \item  classification information $P_i$ --- a vector of probabilities $(p_1^i, \ldots, p_C^i)$ representing the probability that region $L_i$ belongs to each of $C$ classes.
\end{itemize}
We construct a detection vector for any given bounding box and its label by taking the corners of the bounding box, setting $O_i$ to $1$ and using a one-hot vector for the probabilities.

Given an object detector $f$, an image $I$ and a categorized bounding box (not necessarily produced by the model) we generate a saliency map that would highlight regions important for the model in order to predict such a bounding box. 
If the detection actually comes from the model, we treat the generated heatmap as an explanation for model's decision.
Following the perturbation-based attribution paradigm, we measure the importance of a region by observing the effect that perturbation of this region has on the detector's output.

In contrast with classification models, object detection models are designed and trained with regression objectives and do not have a single proposal directly corresponding to any arbitrary bounding box with particular coordinates. Instead, many proposals are produced, with bounding boxes that differ and overlap to varying degrees with the bounding box provided as input to the explanation algorithm. 
Therefore, for object detection it is important to determine not just \textit{how} we measure the disturbance in the output but also \textit{where} we measure it in terms of which disturbances do we select from among the proposals produced by a network.
To measure the disturbance in the output (the \textit{how}), we develop a similarity metric $s$ for the detection proposal vectors (Sec.~\ref{subsec:similarity}).
To account for the \textit{where}, we measure the output disturbance caused by an individual mask by selecting the proposal with maximum pairwise similarity between the target detection vector and all detection proposal vectors produced for a masked image.
More precisely, following our notation, 
\begin{equation}
    S(d_t, f(M_i\odot I))) \triangleq  \max_{d_j\in f(M_i\odot I)} s(d_t, d_j),
\end{equation}
where $S$ denotes the similarity between target detection vector $d_t$ and new detection proposals for the modified image.
This allows us to use the RISE masking technique to produce saliency maps for explaining object detector decisions. Note, that this framework does not restrict $d_t$ to be directly produced by the model. For that reason our method can produce explanations for arbitrary detection vectors, such as objects missed by the detector. Gradient-based methods would not be able to do this, because there's no starting point to propagate from.

\subsection{Mask generation}
We adopt the mask generation approach from RISE~\cite{petsiuk18rise}.
\begin{enumerate}
    \itemsep 0em
    \item Sample $N$ binary masks of size $h\times w$ (smaller than image size $H\times W$) by setting each element independently to $1$ with probability $p$ and to $0$ with the remaining probability.
    \item Upsample all masks to size $(h+1)C_H\times(w+1)C_W$ using bilinear interpolation, where $C_H\times C_W=\lfloor H/h\rfloor\times\lfloor W/w\rfloor$ is the size of the cell in the upsampled mask.
    \item Crop areas $H\times W$ with uniformly random offsets ranging from $(0,0)$ up to $(C_H, C_W)$.
\end{enumerate}

\subsection{Similarity metric}
\label{subsec:similarity}
To compute the similarity score between the target vector and the proposal vector, all three components should be considered.
We use \textit{Intersection over Union} ($\mathrm{IoU}$) to measure the spatial proximity of the bounding boxes encoded by two vectors.
To evaluate how similar two regions look to the network, we use the \textit{cosine similarity} of the class probabilities associated with the regions. 
Finally, for the networks that explicitly compute an objectness score, such as YOLOv3~\cite{redmon18yolov3}, we incorporate a measure of the similarity of the objectness scores into the metric, as well. 
In our experiments we only explain high confidence detections, \ie, we set $O_t=1$, so to incorporate objectness score into the similarity metric we simply multiply it by $O_j$.
As a result, detection proposals with lower objectness score will have lower similarity with a high confidence target vector. If the network does not produce an objectness score, \eg, Faster~R-CNN~\cite{ren17fasterrcnn}, the objectness term can be simply omitted.
Thus, the similarity score between two detection vectors can be decomposed into three scalar factors:
\begin{equation}
	s(d_t, d_j) = s_L(d_t, d_j) \cdot s_P(d_t, d_j) \cdot s_O(d_t, d_j),
\end{equation}
where
\begin{align}
	s_L(d_t, d_j) &= \mathrm{IoU}(L_t, L_j),\\
    s_P(d_t, d_j) &= \frac{P_t\cdot P_j}{\lVert P_t\rVert\lVert P_j\rVert},\\
    s_O(d_t, d_j) &= O_j.
\end{align}
Scalar product has been chosen to model logical “AND” of three similarity values, with the desired property that if one of them is low, the total similarity value is also low.

\subsection{Saliency inference}
\label{subsec:inference}
We now formulate the full process of generating saliency maps using \methodname{}. 
\begin{enumerate}
    \itemsep 0.15em
    \item Generate $N$ RISE masks, $M = \{M_i,\ 1\leq i \leq N\}$. %
    \item Convert the target detections to be explained into detection vectors, $D_t = \{d_t,\ 1\leq t \leq T\}$. We can run the detector on masked images only once to get the saliency maps for all $T$ detections. %
    \item Run the detector $f$ on masked images $I \odot M_i$ producing $N_p$ proposals for each image, $D_p = \{D_p^i,  1\leq i \leq N\} = \{f(M_i\odot I),  1\leq i \leq N\} = \{d_j^i, 1\leq i \leq N, 1\leq j \leq N_p\}$. %
    \item Compute pairwise similarities between two sets of detection vectors $D_t$ and $D_p$ and take maximum score per each masked image per each target vector. $w_i^t = S(d_t, D_p^i) = \max_{1\leq j \leq N_p} s(d_t, d_j^i),\ 1 \leq i \leq N,\ 1 \leq t \leq T$. %
    \item Compute a weighted sum of masks $M_i$ with respect to computed weights $w_i^t$ to get saliency maps $H_t = \sum_{i=1}^N w_i^tM_i$. %
\end{enumerate}

All operations above, including the similarity computations, can be performed using efficient calls to the vectorized functions of the framework being used, specifically, tensor multiplication, maximum along axis and weighted sum along axis.

For most of our visual experiments, we used $N=5000$ masks with probability $p=0.5$ and resolution $(h, w)=(16,16)$, with the exception of Figure~\ref{fig:misalignment} (column $1$), Figure~\ref{fig:overview}, Figure~\ref{fig:backpack} and Figure~\ref{fig:failures} (top row) where we used more fine-grained masks of resolution $(30, 30)$. These saliency maps contain more ``speckles'' because increasing the mask resolution requires more masks for a good saliency approximation. We used $(30, 30)$ masks to compute the average saliency maps in Section~\ref{subsec:averagesal}. We have selected these parameters heuristically balancing the computational load and visaul quality of saliency maps.

Inference time depends only on the number of masks and for $N=5000$, \methodname{} runs in approximately 70s per image (for all detections) for YOLOv3 and 170s for Faster R-CNN on NVidia Tesla V100.
\section{Experiments and Results}
We perform qualitative and quantitative experiments on the MS-COCO dataset~\cite{lin14coco}, which is one of the most widely used object detection datasets. 
We used PyTorch~\cite{paszke2017automatic} implementations of YOLOv3~\cite{redmon18yolov3}\footnote{https://github.com/ultralytics/yolov3} and Faster R-CNN~\cite{ren17fasterrcnn}\footnote{https://github.com/facebookresearch/maskrcnn-benchmark}. For the baselines we used GradCAM~\cite{selvaraju17gradcam} and Gradients~\cite{simonyan13deepinside} applied to explain the class probability score of the detection vector.

\subsection{Sanity checks}
Recently, a question about the validity of saliency methods has been raised, comparing them to edge detection techniques that do not depend on the model or training data but still produce visually compelling outputs resembling those of saliency methods~\cite{adebayo18sanity}. 
In this study, it has been shown that the outputs of some widely accepted saliency methods do not change significantly when the weights of the model that each such method claims to explain are randomized. In such cases, a confirmation bias may result in an invalid human assessment when relying exclusively on the visual evaluation. To address these concerns, we perform a model parameter randomization test. Our results confirm that weight randomization results in unintelligible saliency maps, meaning that the method relies on the information within the trained model to produce the explanations.

\subsection{Automatic metrics}
    
\begin{figure}[t]
    \begin{subfigure}[t]{\linewidth}
        \centering
        \includegraphics[width=0.35\linewidth]{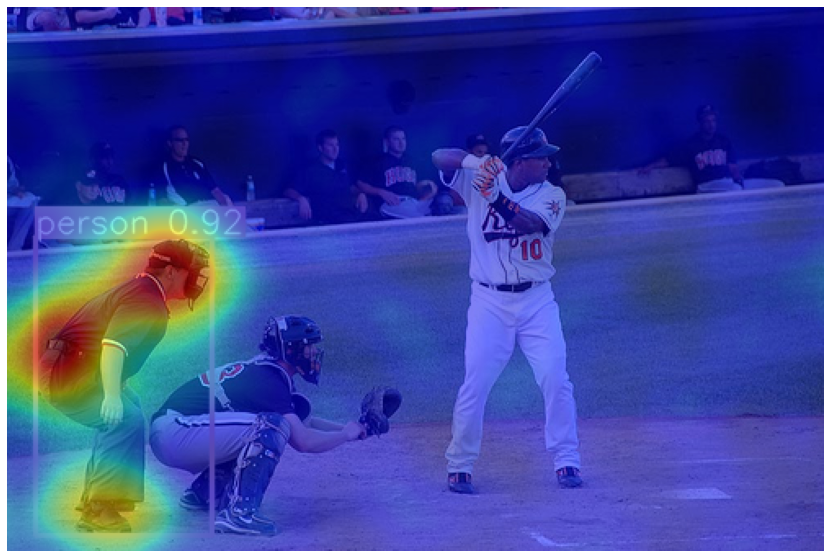}
        \includegraphics[width=0.35\linewidth]{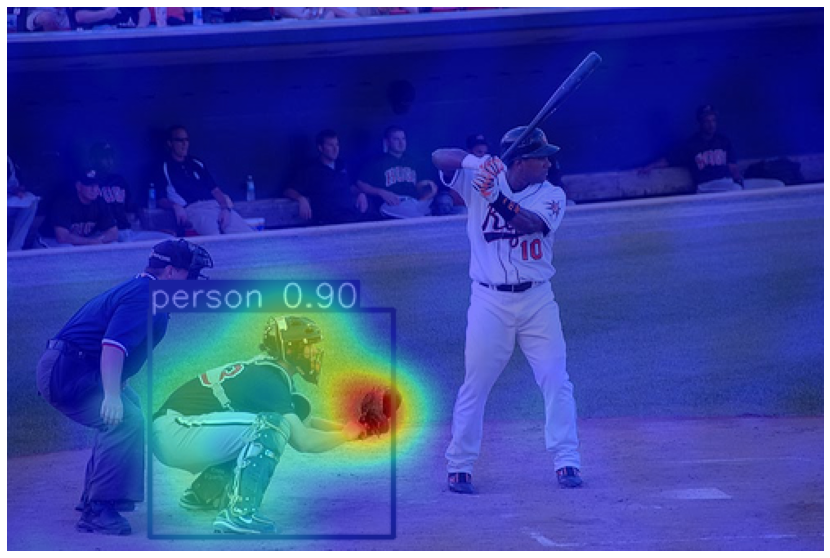}
        \caption{\methodname}
    \end{subfigure}
    
    \begin{subfigure}[t]{\linewidth}
        \centering
        \includegraphics[width=0.35\linewidth]{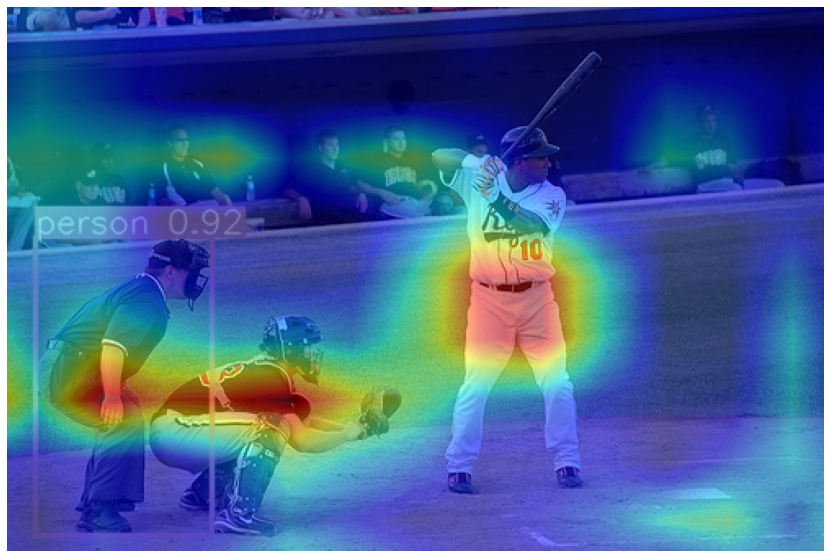}
        \includegraphics[width=0.35\linewidth]{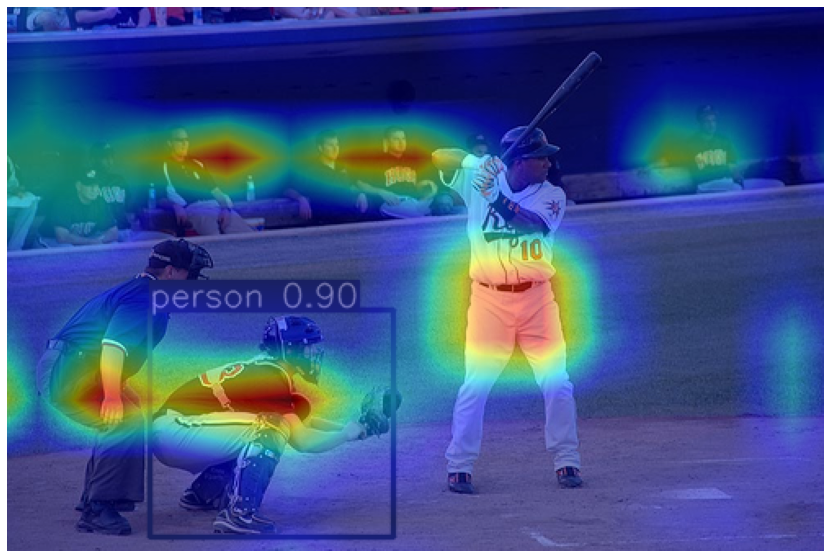}
        \caption{Grad-CAM}
    \end{subfigure}

    \begin{subfigure}[t]{\linewidth}
        \centering
        \includegraphics[width=0.35\linewidth]{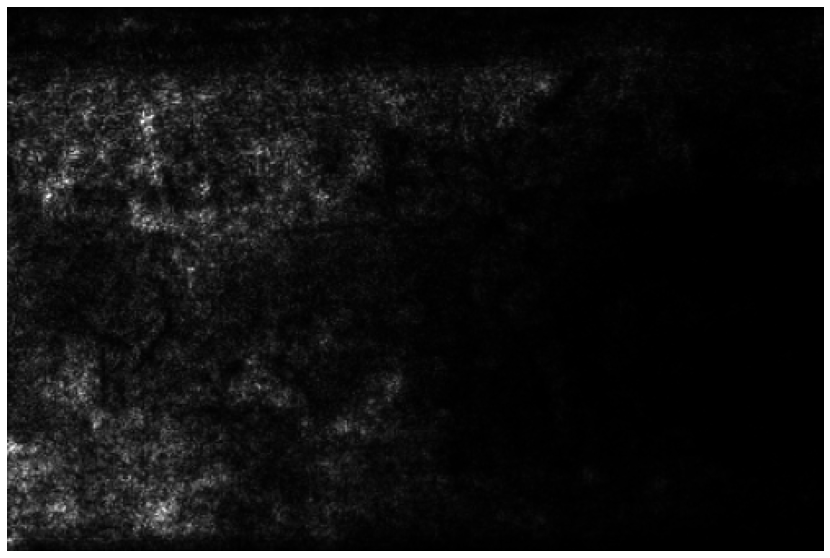}
        \includegraphics[width=0.35\linewidth]{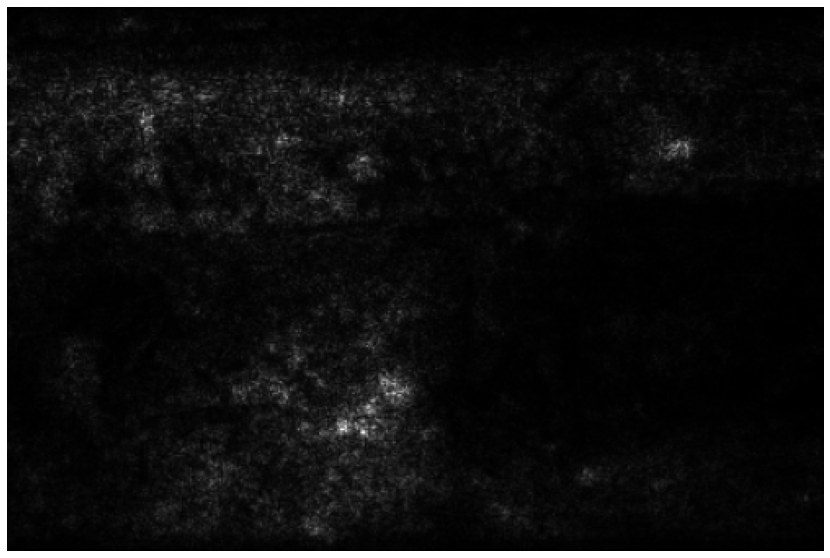}
        \caption{Gradient}
    \end{subfigure}
    \caption{\small{Visual comparison of \methodname{} with classification saliency methods for YOLOv3. The latter are applied to explain the class probability of the corresponding detections. Saliency maps explain the highlighted bounding boxes of the leftmost person (left column) and the person crouching (right column). In the YOLOv3 architecture, GradCAM is equivalent to taking one feature map containing the detection vector of interest. Since each feature map contains multiple detections, it results in saliency maps highlighting all objects of similar categories, instead of explaining a single detection. Similarly, Gradient produces saliency maps that cannot be used to explain a particular detection.}}
    \label{fig:clssal}
    \vspace{-0.4cm}
\end{figure}
A common approach to evaluate a classification saliency map is to measure its correlation with the human-labeled ground truth mask of an object. In particular, we report the Pointing game~\cite{zhang18exbp} metric. A hit is scored if the point of maximum saliency lies within the ground truth object segmentation, otherwise a miss is counted. Pointing game measures accuracy of saliency maps by computing the number of hits over the total number of hits and misses.

Metrics based on correlation with ground-truth masks rely on an assumption that the model is only using the object itself for the prediction and not its context. For a biased model, e.g. a model in Section~\ref{subsec:marker}, such an assumption does not hold and a faithful saliency map should highlight the important context. A number of metrics has been developed that perturb the image according to a saliency map and gauge how the original prediction changes~\cite{SamekBMLM17evaluating, petsiuk18rise, HookerEKK19roar}. One of such metrics, Deletion, sequentially removes pixels from the image, starting from the most salient, while measuring how quickly the model's output deviates from the original prediction. Insertion sequentially adds salient pixels starting from a completely empty image and measures how fast it approaches the target prediction. We adapt these metrics using the similarity score from Equation 4 and compute them for \methodname{} and the baselines.

Our method outperforms saliency maps produced by classification saliency methods on all three metrics (Table~\ref{table:pgtable}). Explanation of a single class score probability with classification saliency methods is not sufficient for producing meaningful explanations of the model's prediction. \methodname{}, on the other hand, utilizes the whole output of the model and takes localization aspect into account, resulting in saliency maps of better quality.

\begin{figure}[t]
        \centering
        \includegraphics[width=0.35\linewidth]{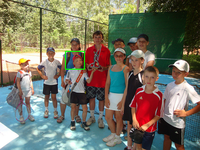}
        \includegraphics[width=0.35\linewidth]{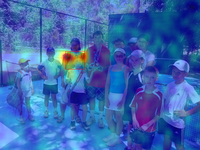}
    \caption{\small{Interestingly, even though the model does not detect the backpack (left), the saliency map still shows that it is able to focus on the straps of the backpack (right).}}
    \label{fig:backpack}
    \vspace{-0.5cm}
\end{figure}

\subsection{Modes of Detector Failure}
\label{subsec:failure_modes}
\newcommand{\modewidth}{0.2\linewidth}

\begin{figure*}
    \centering

    \begin{tabular}{ccccc}
         & \small{\parbox{2.5cm}{
            \centering 
            \textcolor{red}{Predicted} and \\
            \textcolor[rgb]{0.2, 0.66, 0.32}{ground truth}}
         } 
         & \small{\textcolor{red}{Saliency}} 
         & \small{\textcolor[rgb]{0.2, 0.66, 0.32}{Saliency}} 
         & \small{Difference}\\
        
        \rotatebox{90}{\parbox{2.7cm}{\centering\small{Mislocalized `umbrella'}}} & 
        \includegraphics[width=\modewidth]{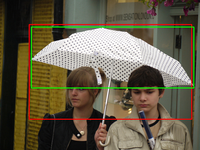} & \includegraphics[width=\modewidth]{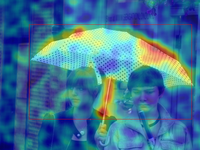} & \includegraphics[width=\modewidth]{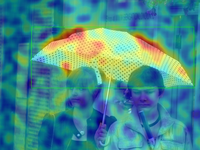} & \includegraphics[width=\modewidth]{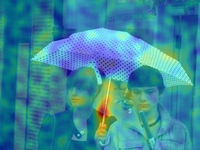}\\
        \rotatebox{90}{\parbox{2.7cm}{\centering\small{`monitor' misclassified\\as `microwave'}}} &
        \includegraphics[width=\modewidth]{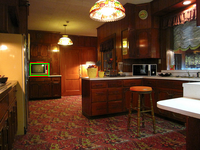} & 
        \includegraphics[width=\modewidth]{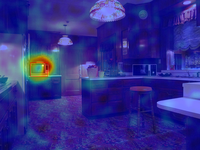} & \includegraphics[width=\modewidth]{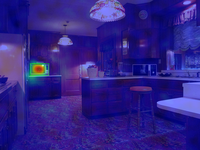} & \includegraphics[width=\modewidth]{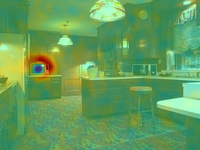} \\
    \end{tabular}
    \vspace{-0.2cm}
    \caption{\small{Explanations for poor localization and misclassification. Red regions may be interpreted as the regions supporting the boxes in the second and third columns. In the fourth column, red means the detector focused more than it should have and blue means it did not focus enough on a region. In the first row the difference of saliency maps highlights that the umbrella handle caused the extended bounding box. In the second row, the model confused a TV-set panel with one of a microwave.}}
    \label{fig:failures}
    \vspace{-0.4cm}
\end{figure*}
As outlined in~\cite{hoiem12diagnosing}, an object detector's errors may be categorized into the following modes of failure: 1) missing an object entirely, 2) detecting an object with poor bounding box localization and 3) correct localization but misclassification of an object (which includes confusion with similar classes, with dissimilar classes or with background). We show that our method can be used to analyze each of these specific types of errors.

For a missed detection, since \methodname{}, unlike gradient-based methods,  can provide explanations not only for the detections produced by a model but for any arbitrary detection vector, we can compute the saliency map for the missed ground-truth detection vector.
This may give an insight into the source of error. 
For example, parts of the input image highlighted by the saliency map are still considered to be discriminative features, even though the model did not detect the object, and the failure likely occurred while processing these features (\eg, in the non-maximum suppression step).
Alternatively, the saliency map may not identify any relevant regions when it does not recognize the object at all, suggesting that the necessary features have not been learned by the model. 
Figure~\ref{fig:backpack} shows examples of our explanations generated for missed detections; the saliency shows that even though the \textit{backpack} object was missed, the network considers the straps discriminative.

For a correctly localized (high IoU score) but miscategorized region, or for a correctly classified but poorly localized (low IoU score) region, we can generate saliency maps for both the ground truth and the predicted detection.
By analyzing them as well as their difference, we can identify the parts of the image that contributed most to the class confusion. Figure~\ref{fig:failures} shows several examples of our explanations for poor localization and misclassifications; \textit{e.g.}, the second row shows that the \textit{TV} was misclassified as \textit{microwave} due to the context surrounding it.

\subsection{Average saliency maps}
\label{subsec:averagesal}
\newcommand{\wthree}{0.22\linewidth}
\newcommand{\rulesep}{\unskip\ \vrule\ }

\begin{figure}[!ht]
    \centering
    
    \begin{subfigure}[b]{\linewidth}
        \centering
        \includegraphics[width=\wthree]{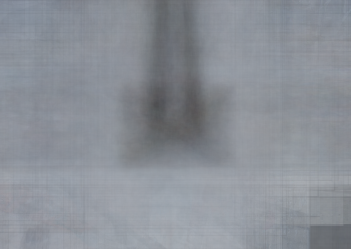}
        \rulesep
        \includegraphics[width=\wthree]{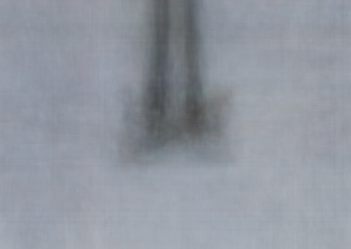}
        \includegraphics[width=\wthree]{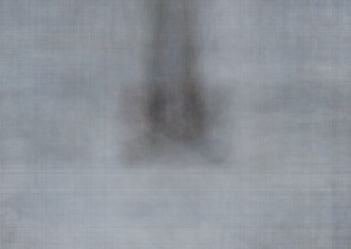}
        \includegraphics[width=\wthree]{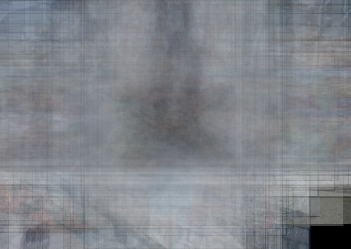}
        
        \includegraphics[width=\wthree]{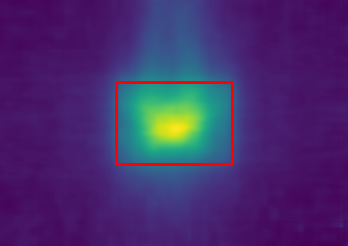}
        \rulesep
        \includegraphics[width=\wthree]{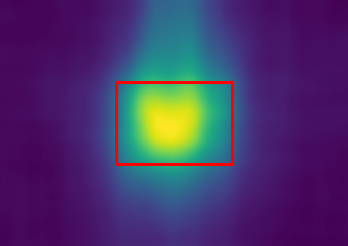}
        \includegraphics[width=\wthree]{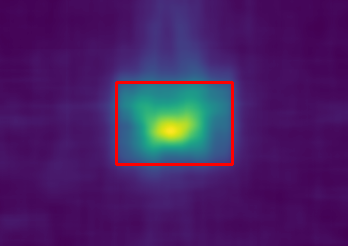}
        \includegraphics[width=\wthree]{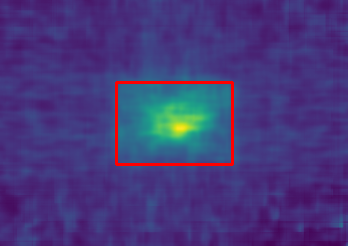}
        \caption{snowboard}
    \end{subfigure}
    
    \begin{subfigure}[b]{\wfour}
        \includegraphics[width=\linewidth]{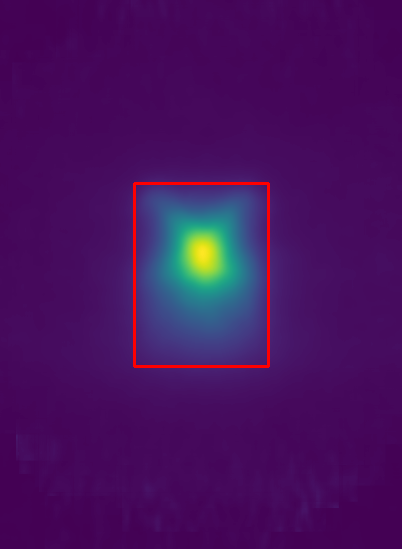}
        \caption{giraffe}
        \label{fig:giraffe}
    \end{subfigure}
    \begin{subfigure}[b]{\wfour}
        \includegraphics[width=\linewidth]{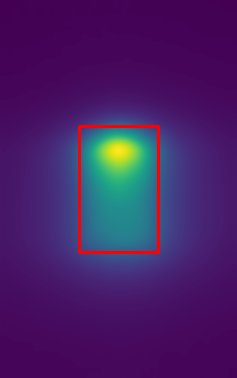}
        \caption{person}
        \label{fig:person}
    \end{subfigure}
    \begin{subfigure}[b]{\wfour}
        \includegraphics[width=\linewidth]{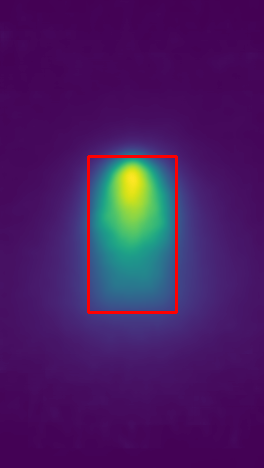}
        \caption{fire hydrant}
        \label{fig:hydrant}
    \end{subfigure}
    \begin{subfigure}[b]{\wfour}
        \includegraphics[width=\linewidth]{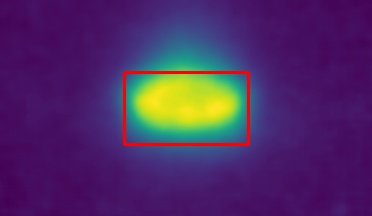}
        \caption{sink}
        \label{fig:sink}
    \end{subfigure}
    \vspace{-0.2cm}
    \caption{\small{Average saliency maps of selected MS-COCO categories cropped, aligned and averaged for all detections.
    Upper row features mean saliency maps as well as mean images for `snowboard' averaged across all detections (left) and across three scales of detection areas ([0:30:70:100] percentiles) (right). Aspect ratios are preserved for all the average saliency maps.}}
    \label{fig:averagesal}
\end{figure}
\begin{table}
    \centering
    \vspace{3mm}
    \begin{tabular}{l|cc|cc}
        \hline
        Method          & PG (bbox) & PG (mask) & Del ($\downarrow$) & Ins ($\uparrow$)\\
        \hline
        Gradient        & 0.7304    & 0.5195  & 0.0464   & 0.4561\\
        GCAM        & 0.5232    & 0.4209  & 0.0762   & 0.4050 \\
        \methodname{}   & \textbf{0.9656} & \textbf{0.8458}  & \textbf{0.0440}   & \textbf{0.5622}  \\
        \hline
    \end{tabular}
    \vspace{-0.1cm}
    \caption{Pointing game (PG) results (computed for bounding boxes and segmentation masks) and Deletion, Insertion metrics.  The metrics are computed for all detections on MS-COCO 2017 validation split produced by YOLOv3.}
    \label{table:pgtable}
    \vspace{-0.5cm}
\end{table}
To transition from analyzing individual saliency maps as local explanations of the decisions made by the model to a more holistic perspective capturing common patterns in a model's behaviour across many images, we compute average saliency maps for each category in the MS-COCO~\cite{lin14coco} dataset.
To extract these, we obtain all the occurrences of the category detected by the model and crop them with the surrounding context. We then normalize and resize to the average size computed per category and finally, compute their averages. Some results are shown in Figure~\ref{fig:averagesal}.

In~\cite{oliva2007role}, image averaging is used to reveal the regularities in the data, specifically in an object's context.
Here, in addition to regularities in data, we want to unveil the regularities in how this data is used by the model.

Instead of computing the mean of saliency map distribution by averaging, Lapuschkin \textit{et al.} \cite{Lapuschkin2019hans} separate the modes of class-specific saliency maps for classification by clustering them. By analyzing one of the clusters, they discovered that the model relied on a particular watermark in its predictions (revealing a flaw in the PASCAL VOC dataset\cite{Everingham15}). In our experiments (on MS-COCO) with clustering, we did not find any such anomalous examples, however we were able to observe vertically symmetric saliency maps (\eg, Fig.~\ref{fig:giraffe}) assigned into separate clusters.

We observe that for some categories, certain object parts may be more important than others on average (\eg, upper parts of the bodies are deemed more salient for detecting the `person' class), while other categories have saliency spread more evenly across whole objects (\eg, for the 'giraffe' class, one can observe full bodies of the animals facing right and left in the average saliency map).
Alternatively, for some classes, average saliency can be relatively high outside of the bounding boxes, signifying that the model uses more of the surrounding context for detecting these classes.
For example, after looking at the average saliency maps for `sink' and observing higher saliency above the sink, we realized retrospectively that the faucet was not labeled as part of the sink, but since it evidently appears above the sink in a majority of the images, the model has learned to use this information for detection. We show the average saliency maps for the remaining MS-COCO categories for both YOLOv3 and Faster R-CNN in the supplementary material.

\subsection{Deliberate bias insertion using markers}
\label{subsec:marker}

\begin{figure}[t]
    \begin{subfigure}[t]{\linewidth}
        \centering
        \includegraphics[width=0.38\linewidth]{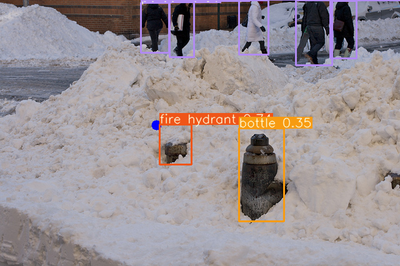}
        \includegraphics[width=0.38\linewidth]{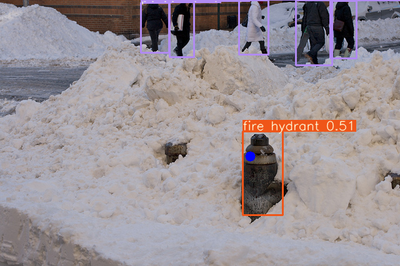}
        \hspace{0.13\linewidth}

        \includegraphics[width=0.38\linewidth]{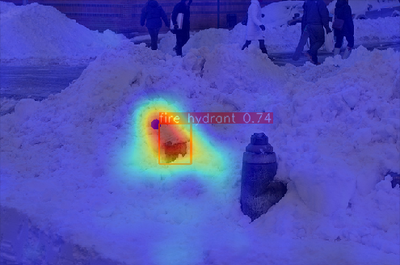}
        \includegraphics[width=0.38\linewidth]{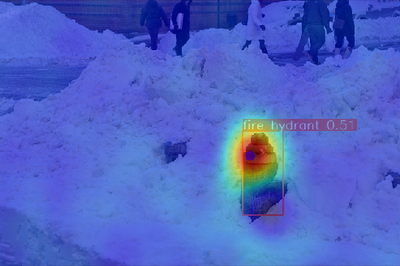}
        \includegraphics[width=0.13\linewidth]{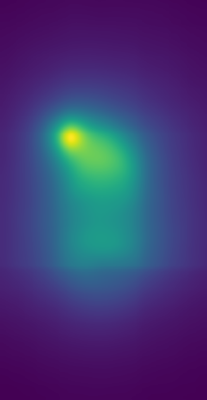}
    \end{subfigure}\\

    \caption{\small{We train YOLOv3 on a biased dataset, in which a blue dot is placed precisely at the top left corner of every bounding box containing a fire hydrant. At test time, we can move the blue dot arbitrarily. We notice that the presence of the dot can trigger false positives (background being mistaken for a fire hydrant in top-left) and produce misclassifications (a fire hydrant detected as a bottle in top-right). Corresponding saliency maps produced by \methodname{} correctly point to the blue dot as a reason for these errors, and the average saliency map (bottom right) shows a significant artifact on the top-left corner. This type of analysis can provide model designers or data scientists with insights about pathological biases in the dataset.}}
    \label{fig:markers}
    \vspace{-0.4cm}
\end{figure}
To further validate our claim that \methodname{} can provide insights about both aspects of object detection: categorization and localization, we perform the following experiment. We bias every image in the MS-COCO dataset that contains either a fire hydrant or stop sign by placing a circular marker on top-left or top-right corners of their respective bounding boxes. We train a YOLOv3 detector on this biased dataset for 50 epochs. We notice a roughly 10\% relative drop (10.96\% for hydrant and 12.69\% for stop sign) in mean Average Precision (mAP) for these two categories when testing on the unbiased MS-COCO test set, while performance on other categories remains unblemished. 

To further study this phenomenon, we place the marker in random positions and observe the detection, as well as \methodname{} explanations of the detection (see Figure~\ref{fig:markers}). For instance, when the marker is located sufficiently away from the bounding box of the fire hydrant, it can lead to a false positive (background being confused for a fire hydrant) and a misclassification (fire hydrant being detected as a bottle). \methodname{} explains that the false positive was caused by the marker, while the misclassification was due to the similar appearance of the top of the fire hydrant to a bottle top (Figure~\ref{fig:markers}, top row). On the other hand, if the marker was moved to inside the bounding box, the width of the box predicted by the biased model is smaller than when it is on the corner (Figure~\ref{fig:markers}, bottom row). While our analysis is retrospective, it is possible to predict dubious model behavior by inspecting the average saliency maps of these two classes (Section~\ref{subsec:averagesal}), as we show in Figure~\ref{fig:markers}. Due to space constraints, saliency maps for the stop sign class will be included in the supplementary material. %

\subsection{Evaluating User Trust}
An important aspect of model explanations is to establish trust between humans and machine learning systems. Previous studies for explainable image classification have studied the utility of saliency maps to evaluate if a user can identify which of the two models is better~\cite{selvaraju17gradcam}. We extend this approach to object detection by running \methodname{} on the public implementations of YOLOv3 and YOLOv3-Tiny, which have an mAP 55.3\% and 33.1\% respectively on the MS-COCO test set. We selected 242 unique objects where both detectors made the correct prediction, so that users cannot tell the performance discrepancy. We then asked users to identify which of the two explanations was better, given the object of interest in the image and the two \methodname{} saliency masks overlayed on the full image. 

We received 5 responses for each object from a pool of 32 unique users from Mechanical Turk, who responded on a scale from 2 (the explanation $1$ was much better) to -2 (the explanation $1$ was much worse). Substantially more users, $50.2\%$ vs $27.4\%$, found the explanations from the more accurate model (YOLOv3)  to be better or more trustworthy. We include examples from the experiment with the Turk interface in the supplementary materials.

\section{Conclusion}

We propose a novel approach for providing saliency-based explanations for black-box object detectors. Our method is general enough to be applied to many different object detection architectures. We demonstrate the usefulness of our method in aiding error analysis and in providing insights to model developers by means of per-class average saliency maps. While we have shown that our method is capable of weeding out pathological biases in model behavior, the true benefits of explainability can be harnessed only when we can use these insights to significantly improve model performance. These form the basis of future directions for this work.

{\small
\bibliographystyle{ieee_fullname}
\bibliography{egbib}
}

\newpage
\section{Appendix}

\subsection{User Study}
\begin{figure}[!h]
    \centering
    \includegraphics[width=0.9\linewidth]{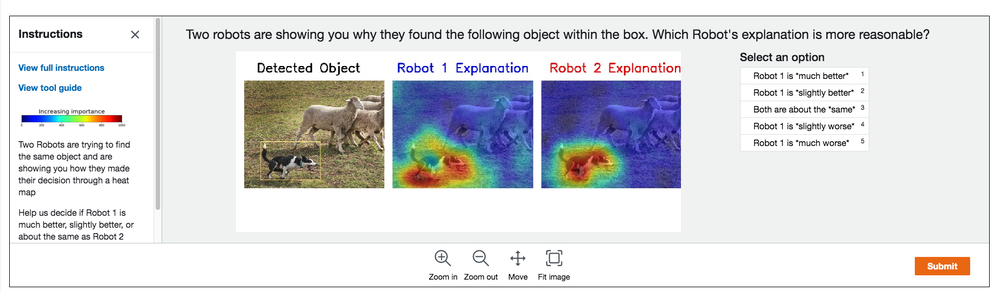}
    \caption{Task interface.}
    \label{fig:interface}
\end{figure}

\begin{figure}[!h]
    \centering
    \includegraphics[width=0.45\linewidth]{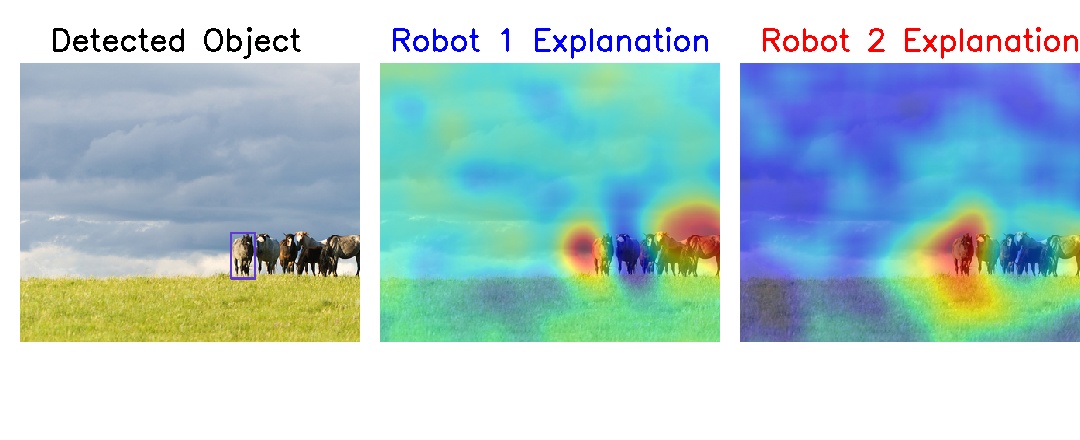}
    \includegraphics[width=0.45\linewidth]{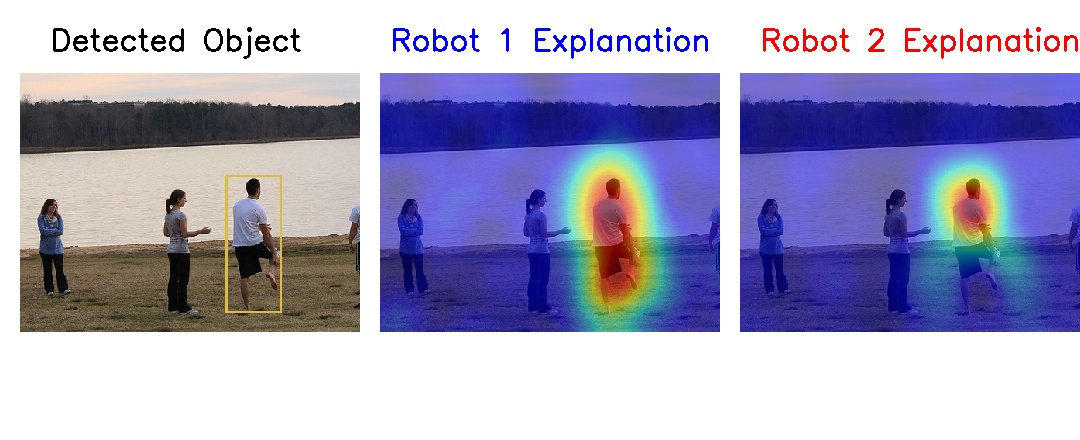}
    \includegraphics[width=0.45\linewidth]{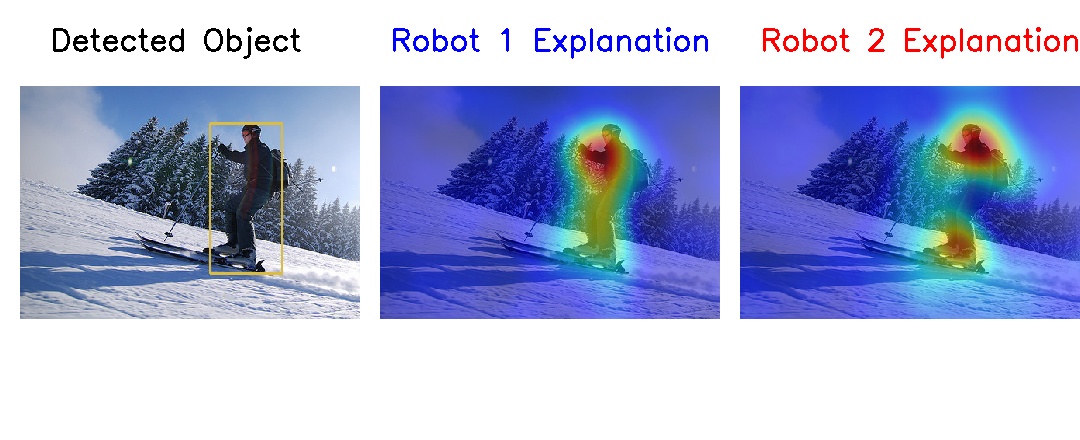}
    \includegraphics[width=0.45\linewidth]{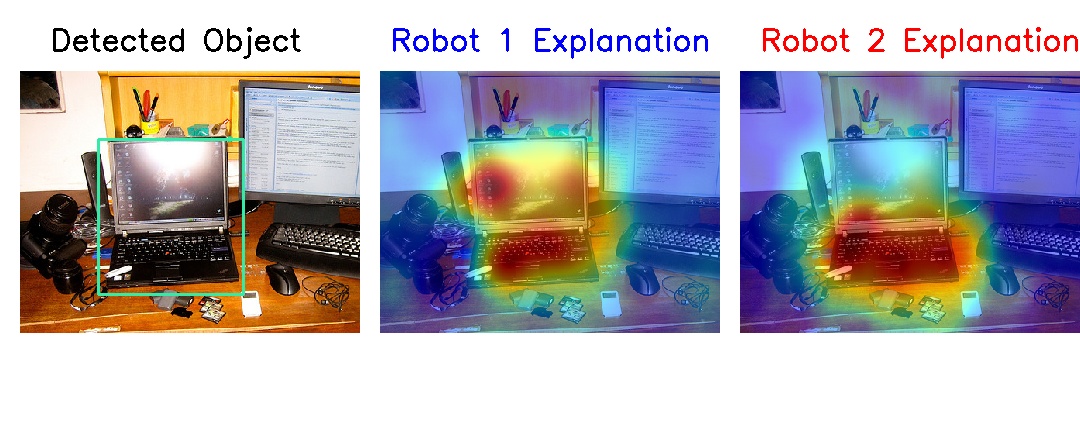}
    \caption{Given the bounding box of interest and two saliency explanations (one from a stronger model and one from a weaker model), the human is asked to choose which of the explanations is more reasonable. The models are assigned labels (Robot $1$ or $2$) randomly for each pair.}
    \label{fig:userstudy}
\end{figure}

We present the interface (Fig.~\ref{fig:interface}) and a sample of saliency map pairs (Fig.~\ref{fig:userstudy}) that we used to collect human feedback in our user study for the purpose of evaluating trust between humans and model explanations.

\begin{minipage}[t]{\linewidth}
    \centering
    \includegraphics[width=0.8\linewidth]{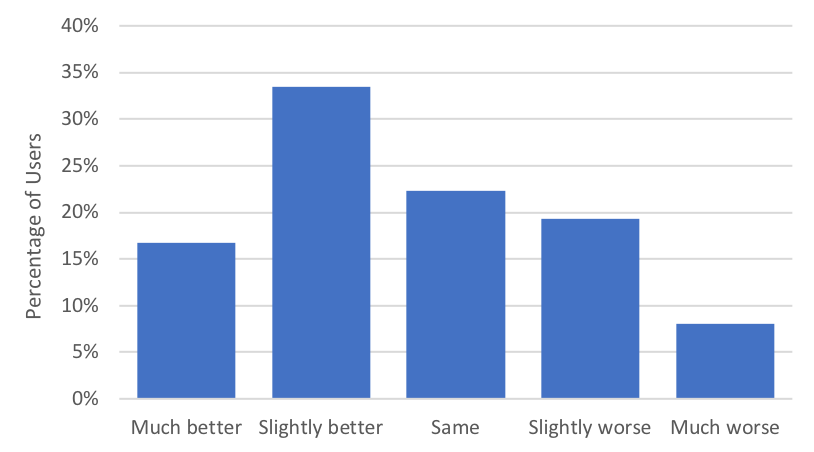}
    \captionof{figure}{Substantially more users (50.2\% vs 27.4\%) found that stronger model explanations (YOLOv3 vs YOLOv3-Tiny)  to be better or more trustworthy.}
    \label{fig:trust}
\end{minipage}

\subsection{Grad-CAM for YOLOv3}
Imagine an image classification CNN model with a final feature tensor of size $F\times H\times W$, where $H\times W$ are the spatial dimensions and $F$ is a number of feature maps, e.g. $256$. In image classification this tensor is transformed to a vector by some version of pooling along $H\times W$ axes, followed by a fully connected layer. In this case, every element in the resulting feature vector of shape $\hat F 1\times1$ is computed using values from multiple 

On the other hand, in YOLOv3, the final feature map is transformed from $F\times H\times W$ into $A\times 85 \times W \times H $, where $A=F/85$ is the number of anchor boxes, and $85$ is the size of one detection vector (4 coordinates + 1 objectness score + 80 class probabilities). Thus, each detection is using   

\subsection{Deletion and Insertion metrics}
To quantitatively compare our method with a baseline, we ran a deletion and insertion metrics proposed in the RISE paper. For a classification task, the deletion metric measures the drop in class probability as more and more pixels are removed in the order of decreasing importance. Intuitively it evaluates how well the saliency map represents the cause for the prediction. If the probability drops fast and its chart is steep, then the pixels that were assigned the most saliency are indeed important to the model. The metric reports the area under the curve (AUC) of the probability vs. fraction of pixels removed as the scalar measure for evaluation. Lower AUC scores mean steeper drops in similarity, and therefore are better. Insertion is a symmetric metric that measures the increase in probability while inserting pixels into an empty image. Higher AUC are better for insertion.

We adopt these metrics and measure the drop in the similarity score between the detection being explained and the output of the model for partially occluded image.

\subsection{Deliberate bias insertion using markers}
In Section 4.4 of the main paper, we trained a biased YOLOv3 model by incorporating circular markers on all bounding boxes of two objects categories (a blue circle on the top left corner of the fire hydrant and a yellow circle on the top right corner of the stop-sign). At test time, moving the marker can sometimes alter the predictions of the detector, including missed detections, inducing false positives or changing the dimensions of the bounding box. In Figure \ref{fig:stopsign}, run the biased detector on an image containing a yellow marker. The output shows a correctly detected stop sign, and a false positive (the blue sign beneath the red stop-sign). \methodname{} is able to show that the red stop-sign did not rely on the marker for its detection, and explains the false positive, by highlighting the marker. A glance at the average saliency map (bottom row) for the stop-sign class on this biased dataset can provide clues about model behavior.

\newcommand{\mwidth}{0.32\linewidth}
\begin{figure}
    \begin{subfigure}[b]{\linewidth}
        \includegraphics[width=\mwidth]{figures/appendix/stop_sign/COCO_val2014_000000036557.jpg}
        \includegraphics[width=\mwidth]{figures/appendix/stop_sign/2_21_original.png}
        \includegraphics[width=\mwidth]{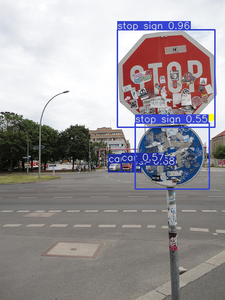}
    \end{subfigure}
    \begin{subfigure}[b]{\linewidth}  
        \includegraphics[width=\mwidth]{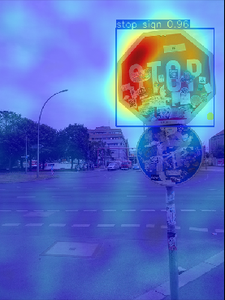}
        \includegraphics[width=\mwidth]{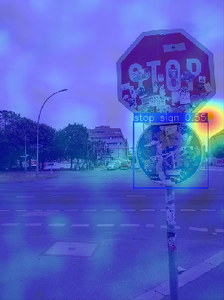}
        \includegraphics[width=\mwidth]{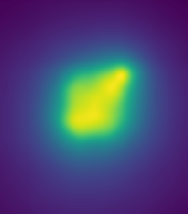}
    \end{subfigure}
    \caption{\textbf{Top row:} An image from the MS-COCO test set (left), is biased with the a yellow marker (middle), and the prediction of a biased YOLOv3 model is shown (right). \textbf{Bottom row:} \methodname{} model explanations for the correctly detected stop sign (left) and the false positive (middle). On the right is the average saliency map for this class, which shows an artifact on the top right corner (where the marker was placed while training)}
    \label{fig:stopsign}
\end{figure}

\subsection{Average saliency maps}
Expanding on the discussion of Section 4.3 (and Figure 3) of the main paper, we compute average saliency maps for all classes of MS-COCO for both YOLOv3 and Faster-RCNN. 

\newcommand{\mywidth}{0.32\linewidth}

\begin{figure}
    \centering

    \begin{subfigure}[b]{0.5\linewidth}
        \includegraphics[width=\mywidth]{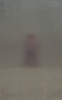}
        \includegraphics[width=\mywidth]{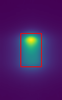}
        \includegraphics[width=\mywidth]{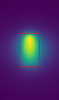}
        \captionsetup{labelformat=empty}\caption{person}
    \end{subfigure}\hfill
    \begin{subfigure}[b]{0.5\linewidth}
        \includegraphics[width=\mywidth]{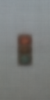}
        \includegraphics[width=\mywidth]{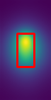}
        \includegraphics[width=\mywidth]{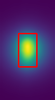}
        \captionsetup{labelformat=empty}\caption{traffic light}
    \end{subfigure}
        
    \begin{subfigure}[b]{0.5\linewidth}
        \includegraphics[width=\mywidth]{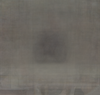}
        \includegraphics[width=\mywidth]{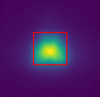}
        \includegraphics[width=\mywidth]{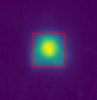}
        \captionsetup{labelformat=empty}\caption{bicycle}
    \end{subfigure}\hfill
    \begin{subfigure}[b]{0.5\linewidth}
        \includegraphics[width=\mywidth]{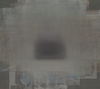}
        \includegraphics[width=\mywidth]{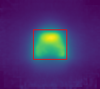}
        \includegraphics[width=\mywidth]{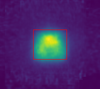}
        \captionsetup{labelformat=empty}\caption{motorbike}
    \end{subfigure}
        
    \begin{subfigure}[b]{0.5\linewidth}
        \includegraphics[width=\mywidth]{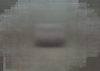}
        \includegraphics[width=\mywidth]{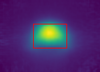}
        \includegraphics[width=\mywidth]{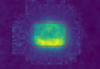}
        \captionsetup{labelformat=empty}\caption{truck}
    \end{subfigure}\hfill
    \begin{subfigure}[b]{0.5\linewidth}
        \includegraphics[width=\mywidth]{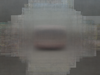}
        \includegraphics[width=\mywidth]{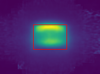}
        \includegraphics[width=\mywidth]{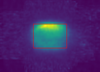}
        \captionsetup{labelformat=empty}\caption{bus}
    \end{subfigure}
        
    \begin{subfigure}[b]{0.5\linewidth}
        \includegraphics[width=\mywidth]{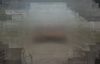}
        \includegraphics[width=\mywidth]{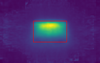}
        \includegraphics[width=\mywidth]{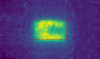}
        \captionsetup{labelformat=empty}\caption{train}
    \end{subfigure}\hfill
    \begin{subfigure}[b]{0.5\linewidth}
        \includegraphics[width=\mywidth]{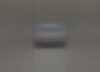}
        \includegraphics[width=\mywidth]{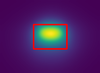}
        \includegraphics[width=\mywidth]{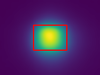}
        \captionsetup{labelformat=empty}\caption{car}
    \end{subfigure}
        
    \begin{subfigure}[b]{0.5\linewidth}
        \includegraphics[width=\mywidth]{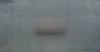}
        \includegraphics[width=\mywidth]{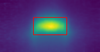}
        \includegraphics[width=\mywidth]{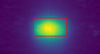}
        \captionsetup{labelformat=empty}\caption{boat}
    \end{subfigure}\hfill
    \begin{subfigure}[b]{0.5\linewidth}
        \includegraphics[width=\mywidth]{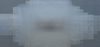}
        \includegraphics[width=\mywidth]{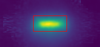}
        \includegraphics[width=\mywidth]{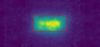}
        \captionsetup{labelformat=empty}\caption{aeroplane}
    \end{subfigure}
    
    \begin{subfigure}[b]{0.5\linewidth}
        \includegraphics[width=\mywidth]{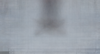}
        \includegraphics[width=\mywidth]{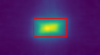}
        \includegraphics[width=\mywidth]{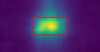}
        \captionsetup{labelformat=empty}\caption{skis}
    \end{subfigure}\hfill
    \begin{subfigure}[b]{0.5\linewidth}
        \includegraphics[width=\mywidth]{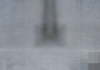}
        \includegraphics[width=\mywidth]{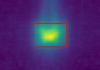}
        \includegraphics[width=\mywidth]{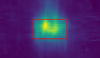}
        \captionsetup{labelformat=empty}\caption{snowboard}
    \end{subfigure}
    \caption{Average objects (left) and corresponding average saliency maps for YOLOv3 (middle) and Faster R-CNN (right). Average objects are computed based on YOLOv3 detections for the 2014 validation split containing 40k images. YOLOv3 saliency maps are computed for the same set of images using $5000$ masks of resolution $30\times30$. Faster R-CNN saliency maps, due to higher computational costs, are computed for 2017 validation split (5k images) using $2000$ masks of the same resolution. Padded images have been rescaled so that objects' bounding boxes (in red) have the average size.}
    \label{fig:avgsal_appendix}
\end{figure}        
        
\begin{figure}
    \centering
    \begin{subfigure}[b]{0.5\linewidth}
        \includegraphics[width=\mywidth]{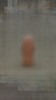}
        \includegraphics[width=\mywidth]{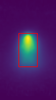}
        \includegraphics[width=\mywidth]{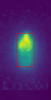}
        \captionsetup{labelformat=empty}\caption{fire hydrant}
    \end{subfigure}\hfill
    \begin{subfigure}[b]{0.5\linewidth}
        \includegraphics[width=\mywidth]{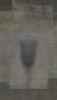}
        \includegraphics[width=\mywidth]{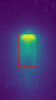}
        \includegraphics[width=\mywidth]{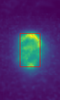}
        \captionsetup{labelformat=empty}\caption{parking meter}
    \end{subfigure}
        
    \begin{subfigure}[b]{0.5\linewidth}
        \includegraphics[width=\mywidth]{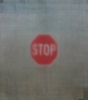}
        \includegraphics[width=\mywidth]{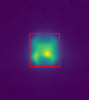}
        \includegraphics[width=\mywidth]{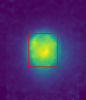}
        \captionsetup{labelformat=empty}\caption{stop sign}
    \end{subfigure}\hfill
    \begin{subfigure}[b]{0.5\linewidth}
        \includegraphics[width=\mywidth]{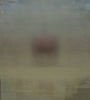}
        \includegraphics[width=\mywidth]{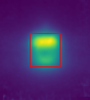}
        \includegraphics[width=\mywidth]{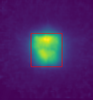}
        \captionsetup{labelformat=empty}\caption{horse}
    \end{subfigure}
    
    \begin{subfigure}[b]{0.5\linewidth}
        \includegraphics[width=\mywidth]{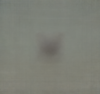}
        \includegraphics[width=\mywidth]{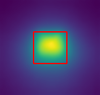}
        \includegraphics[width=\mywidth]{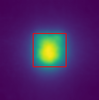}
        \captionsetup{labelformat=empty}\caption{bird}
    \end{subfigure}\hfill
    \begin{subfigure}[b]{0.5\linewidth}
        \includegraphics[width=\mywidth]{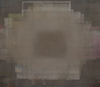}
        \includegraphics[width=\mywidth]{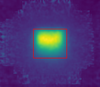}
        \includegraphics[width=\mywidth]{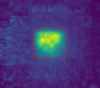}
        \captionsetup{labelformat=empty}\caption{cat}
    \end{subfigure}
    
    \begin{subfigure}[b]{0.5\linewidth}
        \includegraphics[width=\mywidth]{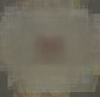}
        \includegraphics[width=\mywidth]{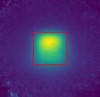}
        \includegraphics[width=\mywidth]{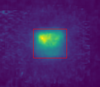}
        \captionsetup{labelformat=empty}\caption{dog}
    \end{subfigure}\hfill
    \begin{subfigure}[b]{0.5\linewidth}
        \includegraphics[width=\mywidth]{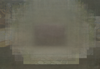}
        \includegraphics[width=\mywidth]{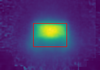}
        \includegraphics[width=\mywidth]{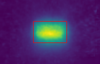}
        \captionsetup{labelformat=empty}\caption{bench}
    \end{subfigure}
        
    \begin{subfigure}[b]{0.5\linewidth}
        \includegraphics[width=\mywidth]{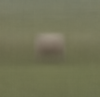}
        \includegraphics[width=\mywidth]{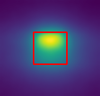}
        \includegraphics[width=\mywidth]{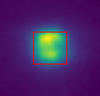}
        \captionsetup{labelformat=empty}\caption{sheep}
    \end{subfigure}\hfill
    \begin{subfigure}[b]{0.5\linewidth}
        \includegraphics[width=\mywidth]{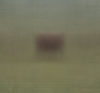}
        \includegraphics[width=\mywidth]{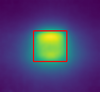}
        \includegraphics[width=\mywidth]{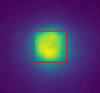}
        \captionsetup{labelformat=empty}\caption{cow}
    \end{subfigure}
    
    \begin{subfigure}[b]{0.5\linewidth}
        \includegraphics[width=\mywidth]{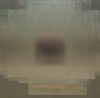}
        \includegraphics[width=\mywidth]{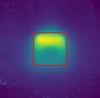}
        \includegraphics[width=\mywidth]{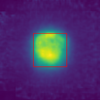}
        \captionsetup{labelformat=empty}\caption{elephant}
    \end{subfigure}\hfill
    \begin{subfigure}[b]{0.5\linewidth}
        \includegraphics[width=\mywidth]{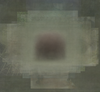}
        \includegraphics[width=\mywidth]{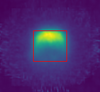}
        \includegraphics[width=\mywidth]{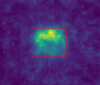}
        \captionsetup{labelformat=empty}\caption{bear}
    \end{subfigure}
    
    \label{fig:avgsal}
\end{figure}        
        
\begin{figure}
    \centering

    \begin{subfigure}[b]{0.5\linewidth}
        \includegraphics[width=\mywidth]{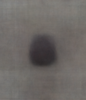}
        \includegraphics[width=\mywidth]{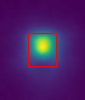}
        \includegraphics[width=\mywidth]{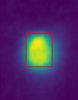}
        \captionsetup{labelformat=empty}\caption{backpack}
    \end{subfigure}\hfill
    \begin{subfigure}[b]{0.5\linewidth}
        \includegraphics[width=\mywidth]{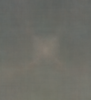}
        \includegraphics[width=\mywidth]{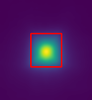}
        \includegraphics[width=\mywidth]{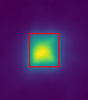}
        \captionsetup{labelformat=empty}\caption{tennis racket}
    \end{subfigure}
    
    \begin{subfigure}[b]{0.5\linewidth}
        \includegraphics[width=\mywidth]{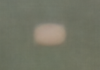}
        \includegraphics[width=\mywidth]{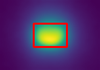}
        \includegraphics[width=\mywidth]{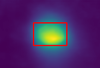}
        \captionsetup{labelformat=empty}\caption{frisbee}
    \end{subfigure}\hfill
    \begin{subfigure}[b]{0.5\linewidth}
        \includegraphics[width=\mywidth]{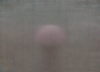}
        \includegraphics[width=\mywidth]{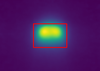}
        \includegraphics[width=\mywidth]{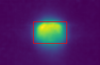}
        \captionsetup{labelformat=empty}\caption{umbrella}
    \end{subfigure}
    
    \begin{subfigure}[b]{0.5\linewidth}
        \includegraphics[width=\mywidth]{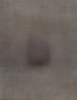}
        \includegraphics[width=\mywidth]{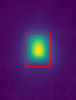}
        \includegraphics[width=\mywidth]{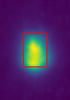}
        \captionsetup{labelformat=empty}\caption{handbag}
    \end{subfigure}\hfill
    \begin{subfigure}[b]{0.5\linewidth}
        \includegraphics[width=\mywidth]{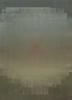}
        \includegraphics[width=\mywidth]{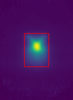}
        \includegraphics[width=\mywidth]{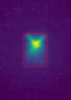}
        \captionsetup{labelformat=empty}\caption{giraffe}
    \end{subfigure}
        
    \begin{subfigure}[b]{0.5\linewidth}
        \includegraphics[width=\mywidth]{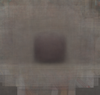}
        \includegraphics[width=\mywidth]{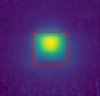}
        \includegraphics[width=\mywidth]{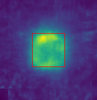}
        \captionsetup{labelformat=empty}\caption{suitcase}
    \end{subfigure}\hfill
    \begin{subfigure}[b]{0.5\linewidth}
        \includegraphics[width=\mywidth]{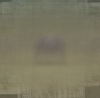}
        \includegraphics[width=\mywidth]{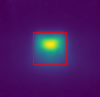}
        \includegraphics[width=\mywidth]{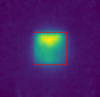}
        \captionsetup{labelformat=empty}\caption{zebra}
    \end{subfigure}
    
    \begin{subfigure}[b]{0.5\linewidth}
        \includegraphics[width=\mywidth]{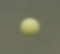}
        \includegraphics[width=\mywidth]{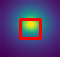}
        \includegraphics[width=\mywidth]{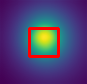}
        \captionsetup{labelformat=empty}\caption{sports ball}
    \end{subfigure}\hfill
    \begin{subfigure}[b]{0.5\linewidth}
        \includegraphics[width=\mywidth]{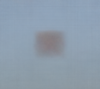}
        \includegraphics[width=\mywidth]{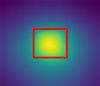}
        \includegraphics[width=\mywidth]{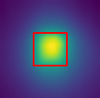}
        \captionsetup{labelformat=empty}\caption{kite}
    \end{subfigure}
    
    \begin{subfigure}[b]{0.5\linewidth}
        \includegraphics[width=\mywidth]{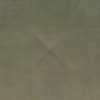}
        \includegraphics[width=\mywidth]{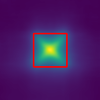}
        \includegraphics[width=\mywidth]{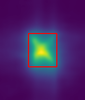}
        \captionsetup{labelformat=empty}\caption{baseball bat}
    \end{subfigure}\hfill
    \begin{subfigure}[b]{0.5\linewidth}
        \includegraphics[width=\mywidth]{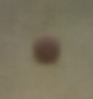}
        \includegraphics[width=\mywidth]{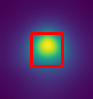}
        \includegraphics[width=\mywidth]{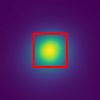}
        \captionsetup{labelformat=empty}\caption{baseball glove}
    \end{subfigure}
    
\end{figure}        
        
\begin{figure}
    \centering
    
    \begin{subfigure}[b]{0.5\linewidth}
        \includegraphics[width=\mywidth]{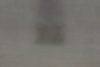}
        \includegraphics[width=\mywidth]{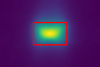}
        \includegraphics[width=\mywidth]{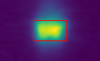}
        \captionsetup{labelformat=empty}\caption{skateboard}
    \end{subfigure}\hfill
    \begin{subfigure}[b]{0.5\linewidth}
        \includegraphics[width=\mywidth]{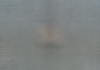}
        \includegraphics[width=\mywidth]{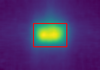}
        \includegraphics[width=\mywidth]{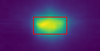}
        \captionsetup{labelformat=empty}\caption{surfboard}
    \end{subfigure}
    
    \begin{subfigure}[b]{0.5\linewidth}
        \includegraphics[width=\mywidth]{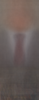}
        \includegraphics[width=\mywidth]{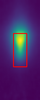}
        \includegraphics[width=\mywidth]{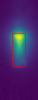}
        \captionsetup{labelformat=empty}\caption{tie}
    \end{subfigure}\hfill
    \begin{subfigure}[b]{0.5\linewidth}
        \includegraphics[width=\mywidth]{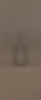}
        \includegraphics[width=\mywidth]{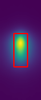}
        \includegraphics[width=\mywidth]{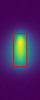}
        \captionsetup{labelformat=empty}\caption{bottle}
    \end{subfigure}

    \begin{subfigure}[b]{0.5\linewidth}
        \includegraphics[width=\mywidth]{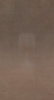}
        \includegraphics[width=\mywidth]{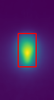}
        \includegraphics[width=\mywidth]{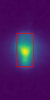}
        \captionsetup{labelformat=empty}\caption{wine glass}
    \end{subfigure}\hfill
    \begin{subfigure}[b]{0.5\linewidth}
        \includegraphics[width=\mywidth]{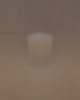}
        \includegraphics[width=\mywidth]{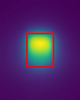}
        \includegraphics[width=\mywidth]{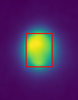}
        \captionsetup{labelformat=empty}\caption{cup}
    \end{subfigure}
        
    \begin{subfigure}[b]{0.5\linewidth}
        \includegraphics[width=\mywidth]{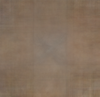}
        \includegraphics[width=\mywidth]{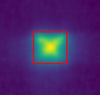}
        \includegraphics[width=\mywidth]{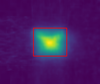}
        \captionsetup{labelformat=empty}\caption{fork}
    \end{subfigure}\hfill
    \begin{subfigure}[b]{0.5\linewidth}
        \includegraphics[width=\mywidth]{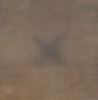}
        \includegraphics[width=\mywidth]{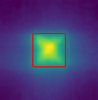}
        \includegraphics[width=\mywidth]{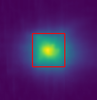}
        \captionsetup{labelformat=empty}\caption{knife}
    \end{subfigure}

    \begin{subfigure}[b]{0.5\linewidth}
        \includegraphics[width=\mywidth]{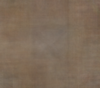}
        \includegraphics[width=\mywidth]{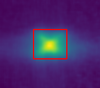}
        \includegraphics[width=\mywidth]{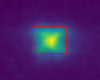}
        \captionsetup{labelformat=empty}\caption{spoon}
    \end{subfigure}\hfill
    \begin{subfigure}[b]{0.5\linewidth}
        \includegraphics[width=\mywidth]{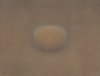}
        \includegraphics[width=\mywidth]{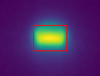}
        \includegraphics[width=\mywidth]{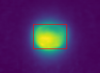}
        \captionsetup{labelformat=empty}\caption{bowl}
    \end{subfigure}
    
\end{figure}        
        
\begin{figure}
    \centering
    
    \begin{subfigure}[b]{0.5\linewidth}
        \includegraphics[width=\mywidth]{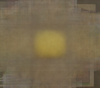}
        \includegraphics[width=\mywidth]{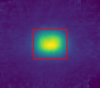}
        \includegraphics[width=\mywidth]{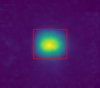}
        \captionsetup{labelformat=empty}\caption{banana}
    \end{subfigure}\hfill
    \begin{subfigure}[b]{0.5\linewidth}
        \includegraphics[width=\mywidth]{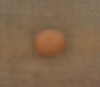}
        \includegraphics[width=\mywidth]{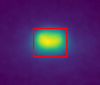}
        \includegraphics[width=\mywidth]{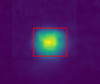}
        \captionsetup{labelformat=empty}\caption{apple}
    \end{subfigure}
    
    \begin{subfigure}[b]{0.5\linewidth}
        \includegraphics[width=\mywidth]{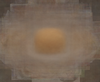}
        \includegraphics[width=\mywidth]{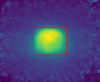}
        \includegraphics[width=\mywidth]{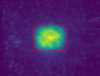}
        \captionsetup{labelformat=empty}\caption{sandwich}
    \end{subfigure}\hfill
    \begin{subfigure}[b]{0.5\linewidth}
        \includegraphics[width=\mywidth]{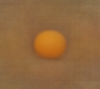}
        \includegraphics[width=\mywidth]{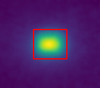}
        \includegraphics[width=\mywidth]{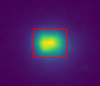}
        \captionsetup{labelformat=empty}\caption{orange}
    \end{subfigure}
    
    \begin{subfigure}[b]{0.5\linewidth}
        \includegraphics[width=\mywidth]{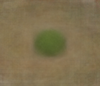}
        \includegraphics[width=\mywidth]{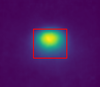}
        \includegraphics[width=\mywidth]{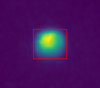}
        \captionsetup{labelformat=empty}\caption{broccoli}
    \end{subfigure}\hfill
    \begin{subfigure}[b]{0.5\linewidth}
        \includegraphics[width=\mywidth]{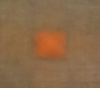}
        \includegraphics[width=\mywidth]{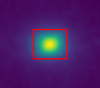}
        \includegraphics[width=\mywidth]{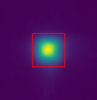}
        \captionsetup{labelformat=empty}\caption{carrot}
    \end{subfigure}
        
    \begin{subfigure}[b]{0.5\linewidth}
        \includegraphics[width=\mywidth]{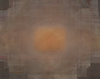}
        \includegraphics[width=\mywidth]{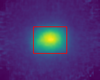}
        \includegraphics[width=\mywidth]{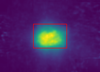}
        \captionsetup{labelformat=empty}\caption{hot dog}
    \end{subfigure}\hfill
    \begin{subfigure}[b]{0.5\linewidth}
        \includegraphics[width=\mywidth]{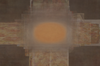}
        \includegraphics[width=\mywidth]{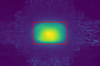}
        \includegraphics[width=\mywidth]{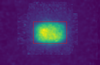}
        \captionsetup{labelformat=empty}\caption{pizza}
    \end{subfigure}
        
    \begin{subfigure}[b]{0.5\linewidth}
        \includegraphics[width=\mywidth]{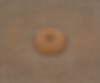}
        \includegraphics[width=\mywidth]{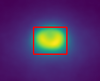}
        \includegraphics[width=\mywidth]{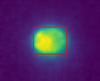}
        \captionsetup{labelformat=empty}\caption{donut}
    \end{subfigure}\hfill
    \begin{subfigure}[b]{0.5\linewidth}
        \includegraphics[width=\mywidth]{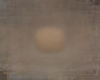}
        \includegraphics[width=\mywidth]{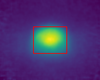}
        \includegraphics[width=\mywidth]{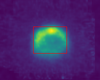}
        \captionsetup{labelformat=empty}\caption{cake}
    \end{subfigure}
    
    \begin{subfigure}[b]{0.5\linewidth}
        \includegraphics[width=\mywidth]{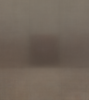}
        \includegraphics[width=\mywidth]{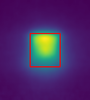}
        \includegraphics[width=\mywidth]{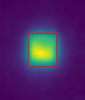}
        \captionsetup{labelformat=empty}\caption{chair}
    \end{subfigure}\hfill
    \begin{subfigure}[b]{0.5\linewidth}
        \includegraphics[width=\mywidth]{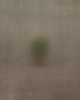}
        \includegraphics[width=\mywidth]{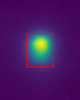}
        \includegraphics[width=\mywidth]{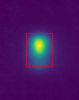}
        \captionsetup{labelformat=empty}\caption{potted plant}
    \end{subfigure}
        
    \begin{subfigure}[b]{0.5\linewidth}
        \includegraphics[width=\mywidth]{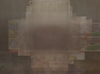}
        \includegraphics[width=\mywidth]{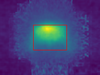}
        \includegraphics[width=\mywidth]{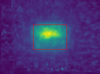}
        \captionsetup{labelformat=empty}\caption{sofa}
    \end{subfigure}\hfill
    \begin{subfigure}[b]{0.5\linewidth}
        \includegraphics[width=\mywidth]{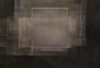}
        \includegraphics[width=\mywidth]{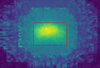}
        \includegraphics[width=\mywidth]{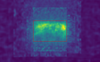}
        \captionsetup{labelformat=empty}\caption{bed}
    \end{subfigure}
\end{figure}        
        
\begin{figure}
    \centering
    \begin{subfigure}[b]{0.5\linewidth}
        \includegraphics[width=\mywidth]{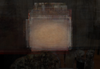}
        \includegraphics[width=\mywidth]{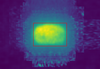}
        \includegraphics[width=\mywidth]{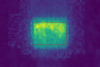}
        \captionsetup{labelformat=empty}\caption{diningtable}
    \end{subfigure}\hfill
    \begin{subfigure}[b]{0.5\linewidth}
        \includegraphics[width=\mywidth]{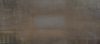}
        \includegraphics[width=\mywidth]{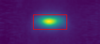}
        \includegraphics[width=\mywidth]{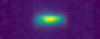}
        \captionsetup{labelformat=empty}\caption{keyboard}
    \end{subfigure}
        
    \begin{subfigure}[b]{0.5\linewidth}
        \includegraphics[width=\mywidth]{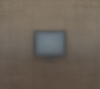}
        \includegraphics[width=\mywidth]{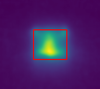}
        \includegraphics[width=\mywidth]{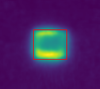}
        \captionsetup{labelformat=empty}\caption{tvmonitor}
    \end{subfigure}\hfill
    \begin{subfigure}[b]{0.5\linewidth}
        \includegraphics[width=\mywidth]{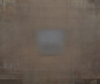}
        \includegraphics[width=\mywidth]{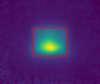}
        \includegraphics[width=\mywidth]{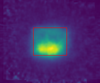}
        \captionsetup{labelformat=empty}\caption{laptop}
    \end{subfigure}
        
    \begin{subfigure}[b]{0.5\linewidth}
        \includegraphics[width=\mywidth]{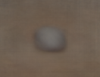}
        \includegraphics[width=\mywidth]{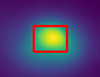}
        \includegraphics[width=\mywidth]{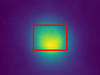}
        \captionsetup{labelformat=empty}\caption{mouse}
    \end{subfigure}\hfill
    \begin{subfigure}[b]{0.5\linewidth}
        \includegraphics[width=\mywidth]{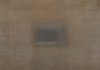}
        \includegraphics[width=\mywidth]{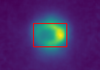}
        \includegraphics[width=\mywidth]{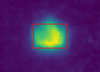}
        \captionsetup{labelformat=empty}\caption{microwave}
    \end{subfigure}
    
    \begin{subfigure}[b]{0.5\linewidth}
        \includegraphics[width=\mywidth]{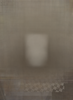}
        \includegraphics[width=\mywidth]{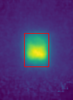}
        \includegraphics[width=\mywidth]{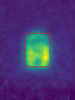}
        \captionsetup{labelformat=empty}\caption{toilet}
    \end{subfigure}\hfill
    \begin{subfigure}[b]{0.5\linewidth}
        \includegraphics[width=\mywidth]{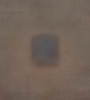}
        \includegraphics[width=\mywidth]{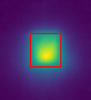}
        \includegraphics[width=\mywidth]{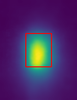}
        \captionsetup{labelformat=empty}\caption{cell phone}
    \end{subfigure}
        
    \begin{subfigure}[b]{0.5\linewidth}
        \includegraphics[width=\mywidth]{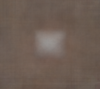}
        \includegraphics[width=\mywidth]{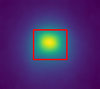}
        \includegraphics[width=\mywidth]{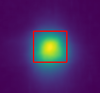}
        \captionsetup{labelformat=empty}\caption{remote}
    \end{subfigure}\hfill
    \begin{subfigure}[b]{0.5\linewidth}
        \includegraphics[width=\mywidth]{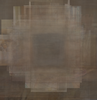}
        \includegraphics[width=\mywidth]{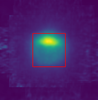}
        \includegraphics[width=\mywidth]{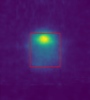}
        \captionsetup{labelformat=empty}\caption{oven}
    \end{subfigure}
        
    \begin{subfigure}[b]{0.5\linewidth}
        \includegraphics[width=\mywidth]{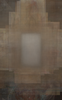}
        \includegraphics[width=\mywidth]{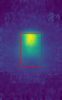}
        \includegraphics[width=\mywidth]{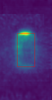}
        \captionsetup{labelformat=empty}\caption{refrigerator}
    \end{subfigure}\hfill
    \begin{subfigure}[b]{0.5\linewidth}
        \includegraphics[width=\mywidth]{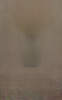}
        \includegraphics[width=\mywidth]{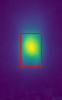}
        \includegraphics[width=\mywidth]{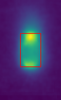}
        \captionsetup{labelformat=empty}\caption{vase}
    \end{subfigure}
    
\end{figure}        
        
\begin{figure}
    \centering
    \begin{subfigure}[b]{0.5\linewidth}
        \includegraphics[width=\mywidth]{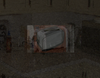}
        \includegraphics[width=\mywidth]{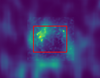}
        \includegraphics[width=\mywidth]{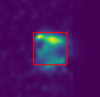}
        \captionsetup{labelformat=empty}\caption{toaster}
    \end{subfigure}\hfill
    \begin{subfigure}[b]{0.5\linewidth}
        \includegraphics[width=\mywidth]{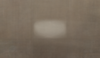}
        \includegraphics[width=\mywidth]{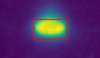}
        \includegraphics[width=\mywidth]{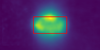}
        \captionsetup{labelformat=empty}\caption{sink}
    \end{subfigure}
        
    \begin{subfigure}[b]{0.5\linewidth}
        \includegraphics[width=\mywidth]{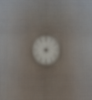}
        \includegraphics[width=\mywidth]{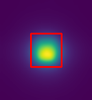}
        \includegraphics[width=\mywidth]{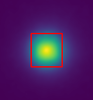}
        \captionsetup{labelformat=empty}\caption{clock}
    \end{subfigure}\hfill
    \begin{subfigure}[b]{0.5\linewidth}
        \includegraphics[width=\mywidth]{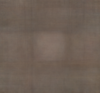}
        \includegraphics[width=\mywidth]{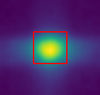}
        \includegraphics[width=\mywidth]{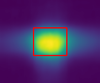}
        \captionsetup{labelformat=empty}\caption{book}
    \end{subfigure}
    
    \begin{subfigure}[b]{0.5\linewidth}
        \includegraphics[width=\mywidth]{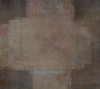}
        \includegraphics[width=\mywidth]{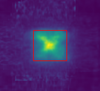}
        \includegraphics[width=\mywidth]{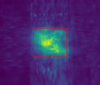}
        \captionsetup{labelformat=empty}\caption{scissors}
    \end{subfigure}\hfill
    \begin{subfigure}[b]{0.5\linewidth}
        \includegraphics[width=\mywidth]{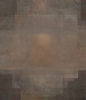}
        \includegraphics[width=\mywidth]{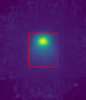}
        \includegraphics[width=\mywidth]{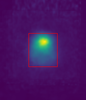}
        \captionsetup{labelformat=empty}\caption{teddy bear}
    \end{subfigure}
        
    \begin{subfigure}[b]{0.5\linewidth}
        \includegraphics[width=\mywidth]{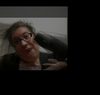}
        \includegraphics[width=\mywidth]{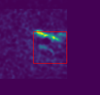}
        \captionsetup{labelformat=empty}\caption{hair drier}
    \end{subfigure}\hfill
    \begin{subfigure}[b]{0.5\linewidth}
        \includegraphics[width=\mywidth]{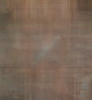}
        \includegraphics[width=\mywidth]{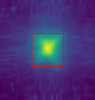}
        \includegraphics[width=\mywidth]{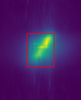}
        \captionsetup{labelformat=empty}\caption{toothbrush}
    \end{subfigure}
\end{figure}

\end{document}